\documentclass{article} %
\usepackage{iclr2023_conference_modified,times}

\usepackage{amsmath,amsfonts,bm}

\def\eqref#1{equation~\ref{#1}}

\def\1{\bm{1}}

\DeclareMathAlphabet{\mathsfit}{\encodingdefault}{\sfdefault}{m}{sl}
\SetMathAlphabet{\mathsfit}{bold}{\encodingdefault}{\sfdefault}{bx}{n}

\usepackage{beton}
\usepackage[T1]{fontenc}
\usepackage{hyperref}
\usepackage{url}
\usepackage{graphicx}
\usepackage{tikz}
\usepackage{comment}
\usepackage{amsmath,amssymb} %
\usepackage{color}
\usepackage{hyperref}
\usepackage{url}
\usepackage[T1]{fontenc}    %
\usepackage{booktabs}       %
\usepackage{amsfonts}       %
\usepackage{nicefrac}       %
\usepackage{microtype}      %
\usepackage{color}
\usepackage{amsmath}
\usepackage{tikz}
\usetikzlibrary{arrows, positioning, automata}
\usepackage{amssymb}
\usepackage{pifont}
\usepackage{graphicx}
\usepackage{multirow}
\usepackage[normalem]{ulem}
\usepackage{enumitem}
\usepackage{wrapfig}
\usepackage{listings}

\definecolor{codegreen}{rgb}{0.36, 0.54, 0.66}
\definecolor{codegray}{rgb}{1,1,1}
\definecolor{codepurple}{rgb}{0.58,0,0.82}
\definecolor{bluepigment}{rgb}{0.2, 0.2, 0.6}
\definecolor{amethyst}{rgb}{0.6, 0.4, 0.8}
\definecolor{backcolour}{rgb}{1,1,1}
\lstdefinestyle{mystyle}{
    backgroundcolor=\color{backcolour},   
    numberstyle=\tiny\color{codegray},
    basicstyle=\ttfamily\tiny,
    breakatwhitespace=false,         
    breaklines=true,                 
    captionpos=b,                    
    keepspaces=true,                 
    numbers=left,                    
    numbersep=5pt,                  
    showspaces=false,                
    showstringspaces=false,
    showtabs=false,                  
    tabsize=2
}
\usepackage{caption}
\usepackage[scr=boondoxo]{mathalfa}
\usepackage[scaled=.8]{beramono}
\usepackage{color, colortbl,booktabs}
\usepackage{algorithm}
\usepackage[noend]{algpseudocode}
\algnewcommand{\LineComment}[1]{\State \(\triangleright\) #1}

\usepackage{graphicx}
\usepackage{amsmath}
\usepackage{amssymb}
\usepackage{booktabs}
\usepackage{subfigure}
\usepackage{pgfplots}
\usepackage{pifont}
\usepackage{caption}
\usepackage{enumitem}
\usepackage{wrapfig}
\usepackage{natbib}
\definecolor{dark-green}{rgb}{0,0.6,0}
\definecolor{amaranth}{rgb}{0.9, 0.17, 0.31}
\definecolor{bostonuniversityred}{rgb}{0.8, 0.0, 0.0}
\definecolor{brightpink}{rgb}{1.0, 0.0, 0.5}
\definecolor{darklava}{rgb}{0.28, 0.24, 0.2}
\definecolor{darkgreen}{rgb}{0.0, 0.2, 0.13}
\definecolor{coolblack}{rgb}{0.0, 0.18, 0.39}
\definecolor{blue-violet}{rgb}{0.54, 0.17, 0.89}

\usepackage{tikz}
\newcommand*\circled[1]{\tikz[baseline=(char.base)]{
            \node[shape=circle,draw,inner sep=0.5pt] (char) {#1};}}

\usepackage{float}
\newfloat{floatquote}{tbp}{loq}
\floatname{floatquote}{Quote}
\usepackage{caption}
\usepackage{comment}

\title{Cross-Reality Re-Rendering: Manipulating between Digital and Physical Realities}
\iclrfinalcopy

\author{Siddhartha Datta
\\
University of Oxford\\
}

\usepackage{cases}
\usepackage{amsmath,amsthm}

\begin{document}
\maketitle

\begin{abstract}
The advent of personalized reality has arrived. Rapid development in AR/MR/VR enables users to augment or diminish their perception of the physical world. Robust tooling for digital interface modification enables users to change how their software operates. As digital realities become an increasingly-impactful aspect of human lives, we investigate the design of a system that enables users to manipulate the perception of both their physical realities and digital realities. Users can inspect their view history from either reality, and generate interventions that can be interoperably rendered cross-reality in real-time. Personalized interventions can be generated with mask, text, and model hooks. Collaboration between users scales the availability of interventions. We verify our implementation against our design requirements with cognitive walkthroughs, personas, and scalability tests.
\end{abstract}

\section{Introduction}
\label{sec:intro}

Cross-reality systems provide users with access to information and objects between multiple realities 
from either reality. 
In prior work in augmented and mixed reality, 
cross-reality systems are methods that transition the level of virtuality along the reality-virtuality continuum \citep{kishino}.
\citet{10.1145/3491102.3501821} demonstrated how an end-user can change the level of virtuality of their physical environment, from one where real objects become virtual objects, to increasing levels of virtuality of the surrounding physical environment, to a completely virtual environment.
Prototypes such as VRoamer \citep{8798074} or ShareVR \citep{10.1145/3025453.3025683} require significant client-side hardware in enabling real-time cross-reality rendering along the continuum.
Physical and virtual realities tend to be the most common target realities to sample objects, information, and scene artifacts.
Head-mounted displays are the primary 
interface
to access cross-reality rendering.

Motivated by how much time humans spend time in both the physical and digital reality,
we investigate how to support the rendering of objects of one reality in the other.
An underlying assumption is that the physical reality and virtual reality have semantic mappings, where the objects from one reality can be mapped to another reality.
Unlike virtual reality,
the digital reality is not viewed through egocentric vision but through heterogeneous graphical user interfaces (e.g. media ranging from text to video, programs ranging from webpages to apps, operating systems ranging from Android to Windows, devices ranging from mobile to desktop).
The view is different, partly due to the modes of interaction in each reality. 
The physical reality requires physical movement (e.g. gaze, body) for a user to actively seek information,
while the digital reality is designed such that information flows to the user with minimal interaction (at most requiring finger action).
The difference in view and affordances contributes to a non-trivial semantic mapping of objects in the digital reality to a physical reality.
When considering alternative reality where this mapping is not provided, transitioning along the continuum between these realities become challenging. 
We are shifting away from transitioning between a physical and semantically-mapped reality.
We investigate how end-users can assist cross-reality rendering by providing semantic information collaboratively.
Other than a semi-supervised approach to constructing mappings between both realities,
enabling end-users to author or reflect on their experiences have brought benefits in augmented \citep{10.1145/3491102.3517665} and digital \citep{10.1145/3479600} realities.

\noindent
\textbf{Contributions}
We
are the first to 
contribute 
a cross-reality rendering system that allows users to manipulate their digital and physical realities
in real-time with respect to each other. 
We are the first to implement an interface-agnostic 
modification framework, 
which modifies digital interfaces
agnostic to operating system, program, or content.
Human-in-the-loop learning enables users to construct interventions, 
and collaboration scales the pool of interventions. 
Along with implementation details,
we validate requirements
with walkthroughs, personas, and scalability tests.

\begin{table*}[t]
\centering
\begin{minipage}{3.5cm}
    \subfigure[\textit{User authentication}: Secure gateway to view history, devices, and camera feed.]{
    \includegraphics[width=2.5cm]{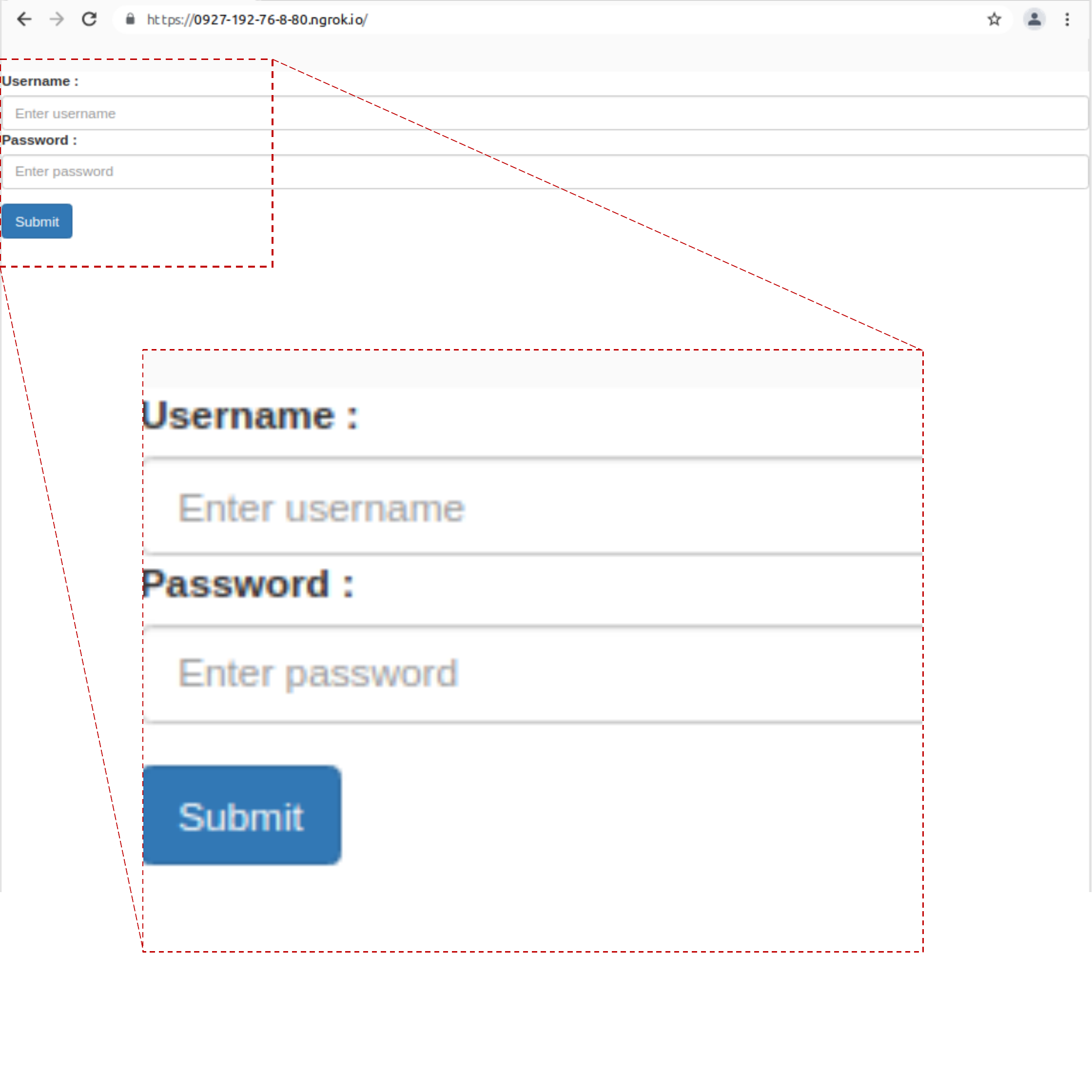}
    } \\
    \subfigure[\textit{Interface \& interventions selection}: Registered devices and available interventions. 
    ]{
    \includegraphics[width=3cm]{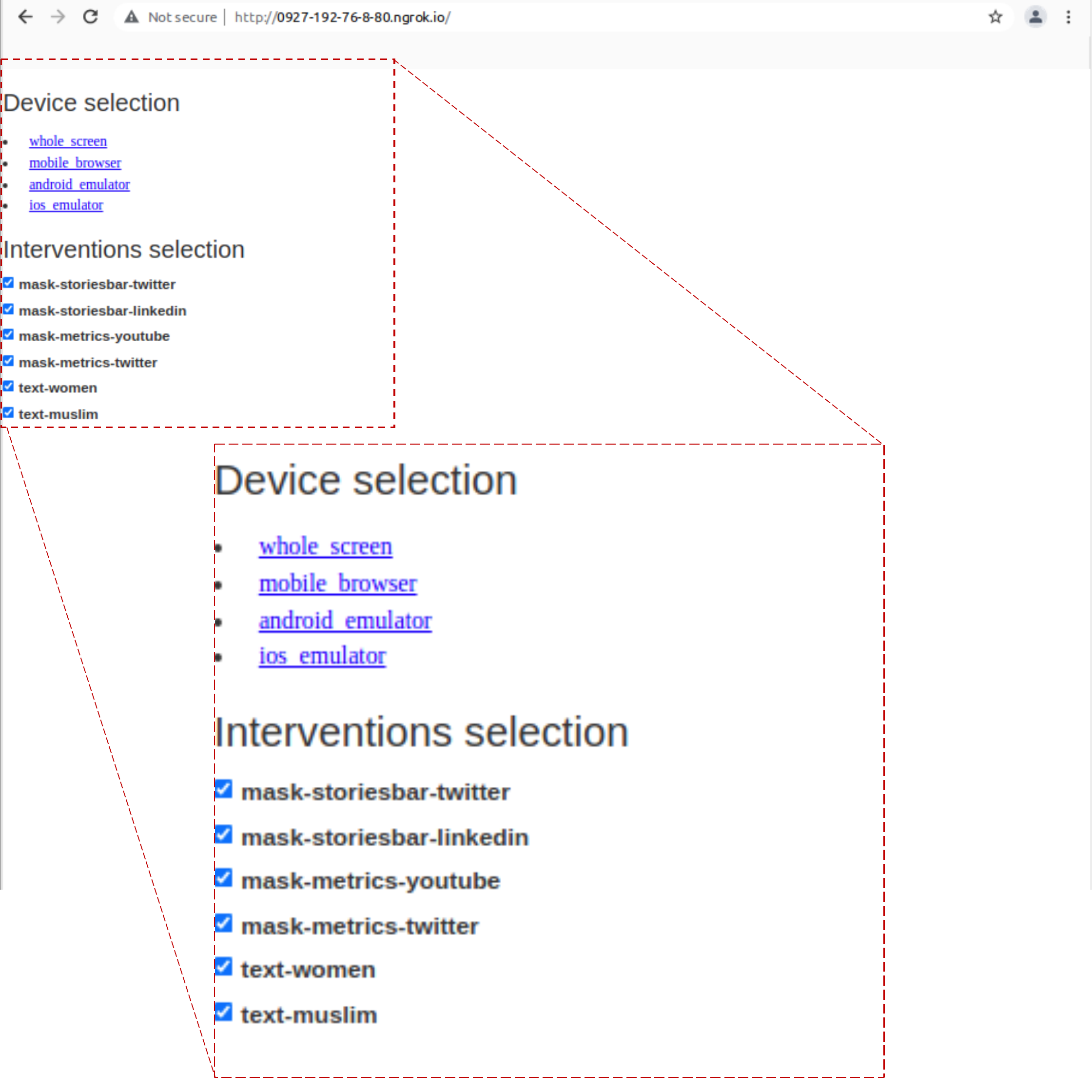}
    }
\end{minipage}
\begin{minipage}{6cm}
    \subfigure[\textit{Interface access}: Accessing a Linux desktop from another (Linux) desktop browser.]{
    \includegraphics[width=6cm]{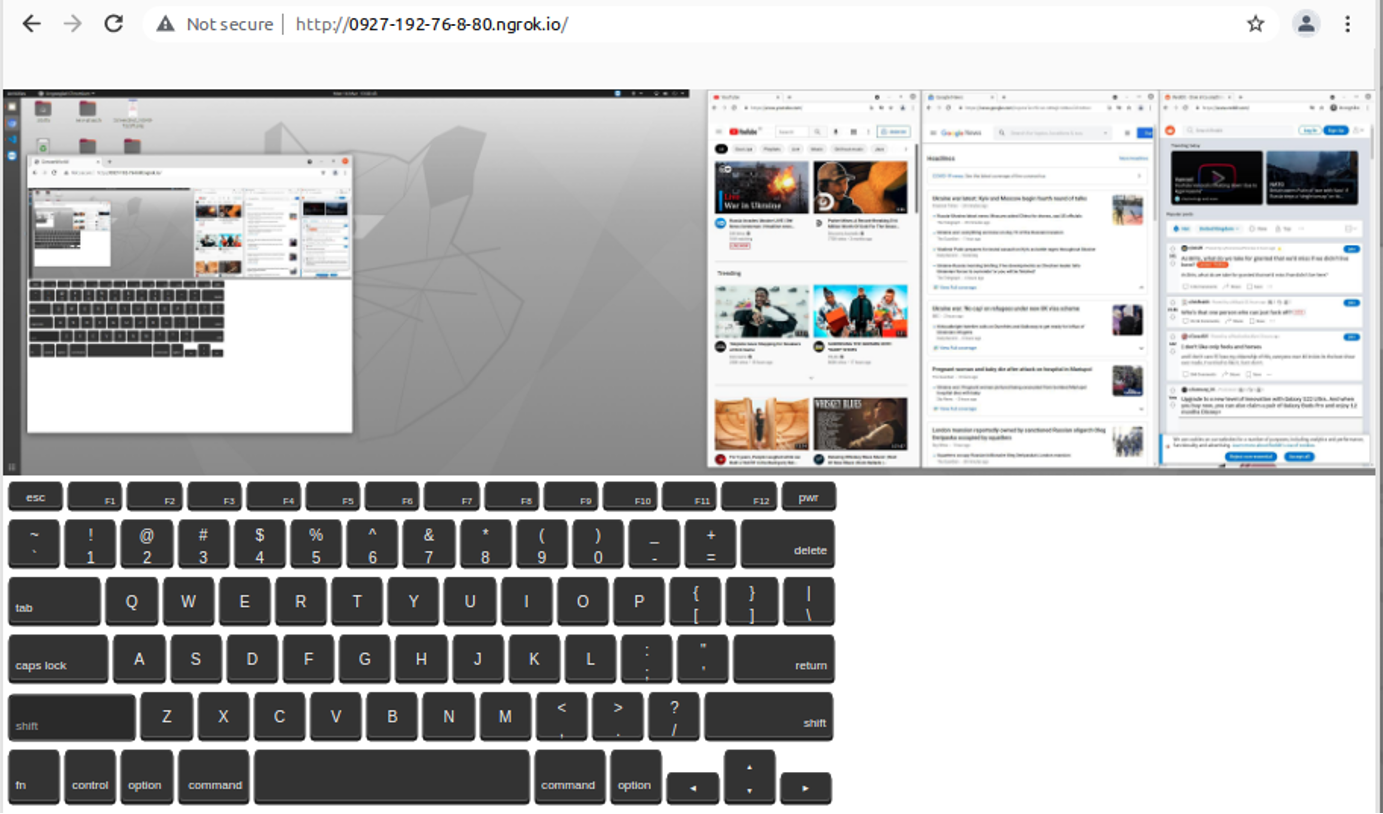}
    } \\
    \subfigure[\textit{Interface access}: Camera feed on the secondary device (to be loaded in full-screen).]{
    \includegraphics[width=6cm]{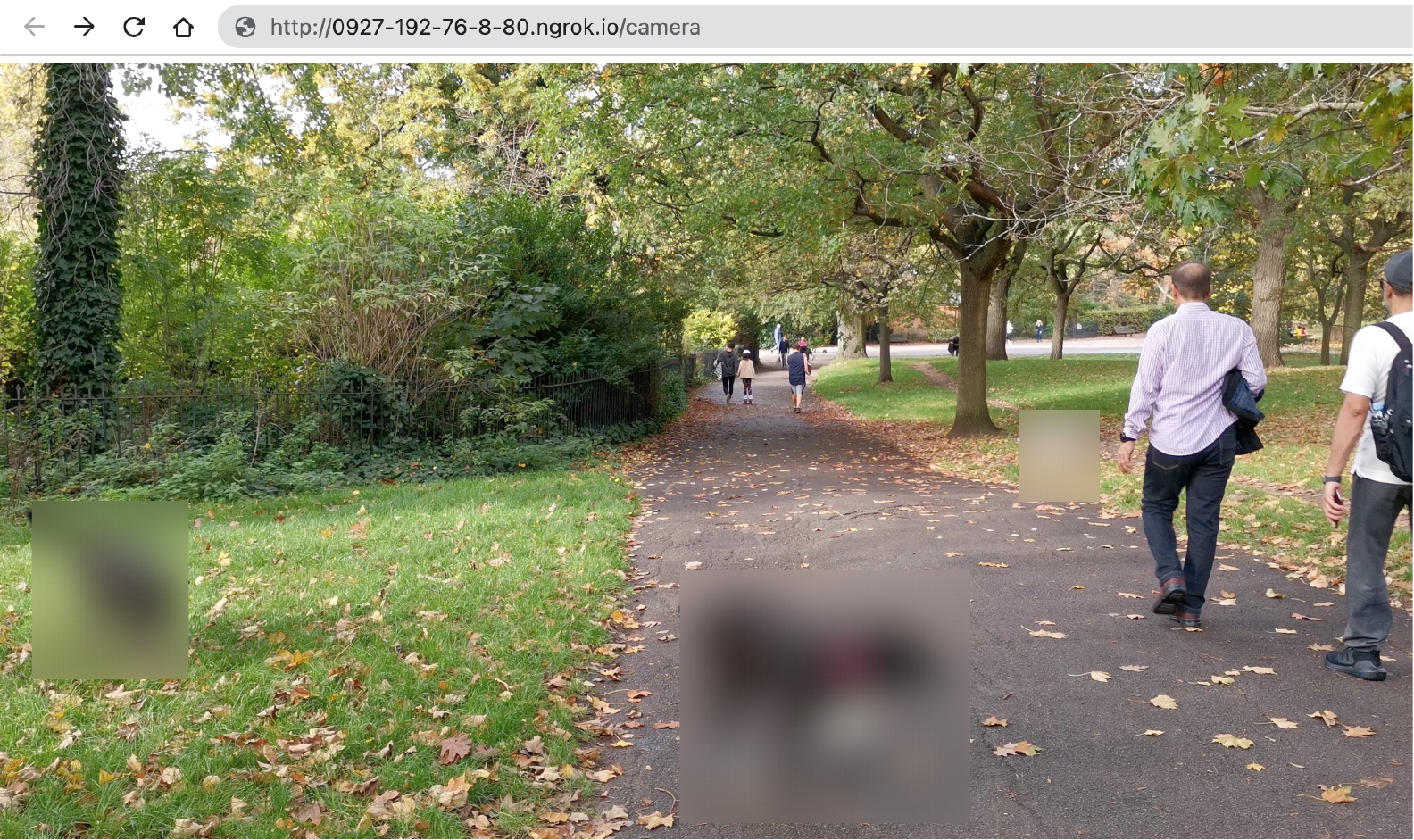}
    }
\end{minipage}
\hspace{0.3cm}
\begin{minipage}{3.8cm}
     \subfigure[\textit{Interface access}: Accessing Android emulator from an Android  device.]{
     \includegraphics[width=3.8cm]{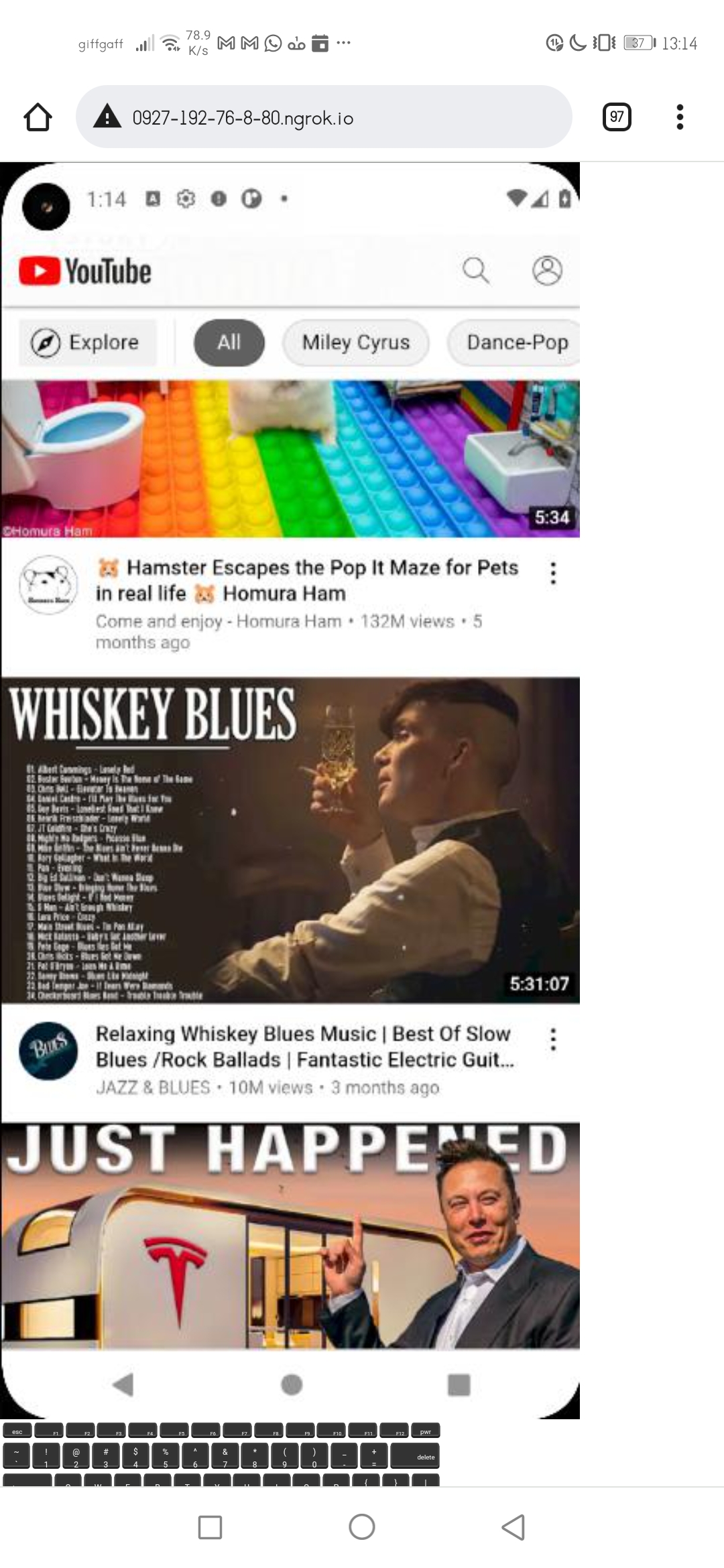}
    }
\end{minipage}
\captionof{figure}{\textit{Walkthrough: }The steps taken by a user to access the different re-rendered realities.}
\label{fig:walkthrough}
\end{table*}
\newpage
\section{Related Work}
\label{sec:background}

\textbf{Distinguishing reality. }
\citet{kishino} elicited the reality–virtuality continuum, 
enabling the interpolation of elements of reality and elements of virtuality. 
We refer a \textit{reality} to the environment in which the user manifests, perceives, and interacts with objects.
\textit{Objects} are manifestations of information.
The \textit{view} refers to the user's point-of-view that allows them to perceive the reality at any given moment (e.g. first-person-view). 
An \textit{interface} is a tool that enables the user to interact with their reality (e.g. device screen, head-mounted display).
The virtual reality has been primarily used as a source of virtual objects, with the intention that these objects can be overlayed onto the physical reality at varying levels.
The digital reality, contrarily, is not a specially-designed reality for objects to be overlayed onto a physical reality (e.g. the digital reality is not geo-spatially organized). 
Given the level of immersion a human has in both physical and digital realities, 
unlike prior work in AR/MR/VR,
we aim to enable users to manipulate both their physical and digital reality, extensibly treating them as a single reality.

\textbf{Manipulating physical reality. }
Rather than immersing in complete reality or virtuality, 
we blend between the two along the continuum.
Two general approaches to the manipulation of real/virtual objects in the physical reality are augmented reality and diminished reality \citep{mann1994}.
Augmented reality adds virtual information onto a reality.
NaviCam \citep{10.1145/215585.215639} was early work demonstrating the placement of messages on video screens.
VRCeption \citep{10.1145/3491102.3501821} enables users to dynamically transition along the reality-virtually continuum.
ScalAR \citep{10.1145/3491102.3517665} enables users to author virtual objects to be placed in their physical reality.
Diminished reality, on the other hand, removes real objects from reality. 
It erases physical objects through 
inpainting \citep{10.1109/ISMAR.2012.6402551},
approximation \citep{10.1145/3126594.3126601},
or multiple cameras \citep{10.1145/3173574.3173703}.
Software from digital interfaces can be ported to be used alongside AR/MR/VR systems (e.g. \citet{horizon}).
While this enables the usage of the software itself,
each reality is still compartmentalized.
For example, what one does on their desktop browsing sessions would play no effect on their physical world viewing experience.

\textbf{Manipulating digital reality. }
The main challenges faced by frameworks that modify digital interfaces are:
interoperability between programs (apps, browsers) and OS, 
requiring escalation of privilege,
significant development/maintenance effort of modifications (e.g. patches break with version changes).
\textit{Code modifications}
make changes to source code,
either installation code to modify software before installation,
or run-time code to modify software during usage
(e.g. browser extensions for desktop/mobile, Cydia Substrate~\citep{cydia} for iOS, Xposed Framework~\citep{xp} for Android).
\textit{External modifications}
require installing a program that affects other programs (e.g. usage tracking with HabitLab~\citep{kovacs_thesis}).
\textit{Overlay modifications} render graphics on an overlay layer over an active interface instance
(e.g. occluding inappropriate text/images with models~\citep{greaseterminator, greasevision}).

\newpage
\section{Cross-Reality System Architecture}

We detail the implementation of the system base, upon which functionality on rendering (Section 4) and playback (Section 5) is supported, 
and further functionality can be added.
All images are captured with our working prototype.

We outline the architectural components as follows:
\begin{enumerate}[noitemsep,topsep=0pt,parsep=0pt,partopsep=0pt]
    \item[] \circled{1} The user logs into the system to access their camera feed or a set of personal emulators, along with a set of interventions.
    The system admin has provisioned a set of emulated devices, hosted on virtual machines on a server.
    The user places a (secondary) hand-held device on their head-mounted display, where the camera feed is streamed to a server and a re-rendered feed is loaded. 
    \item[] \circled{2} The user selects their desired interventions and views re-rendered interfaces.
    \item[] \circled{3} The user accesses their view history and annotates graphics or text for generating interventions, which then re-populate the list of interventions available to members in a network. 
\end{enumerate}

\begin{figure}
    \centering
    \includegraphics[width=0.8\linewidth]{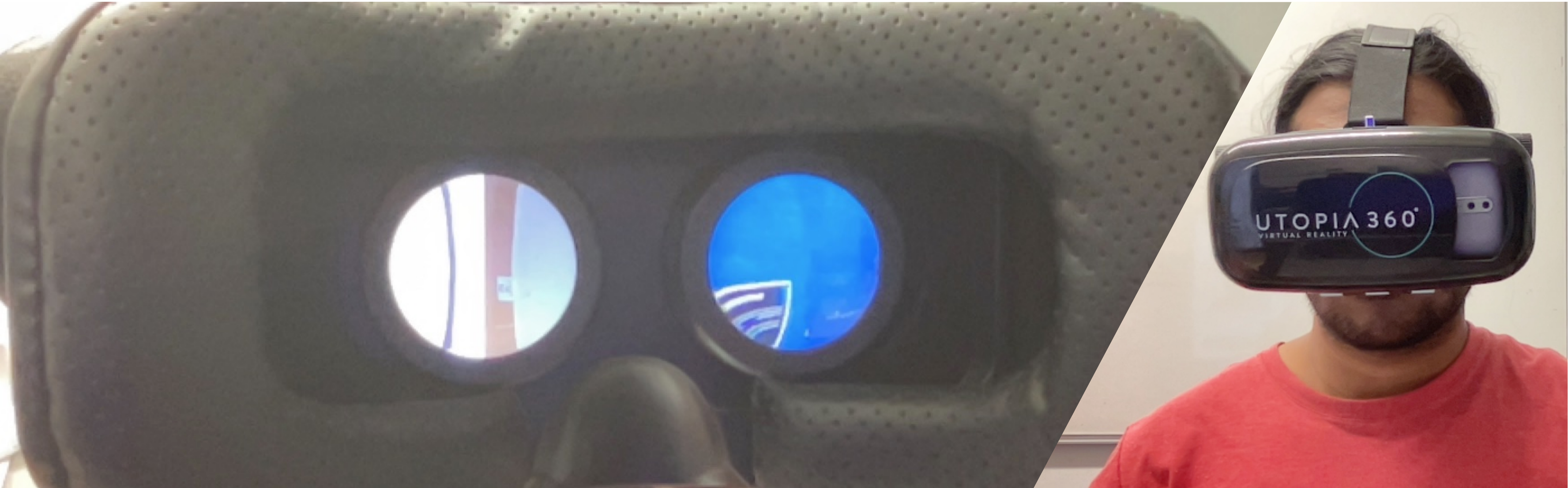}
    \caption{\textit{Head-mounted display}:
    The user's view \textit{(left)}, 
    and the external look \textit{(right)} (with the handheld device inside, its back-facing camera exposed for capturing images).
    }
    \label{fig:headset}
\end{figure}

The user accesses a web application (compatible with desktop/mobile browsers). With their login credentials, the database loads the corresponding mapping of the user's virtual machines that are shown in the \textit{interface selection page} (Figure~\ref{fig:walkthrough}(b)). 
The server carries information on accessing a set of emulated devices.
Each emulator is rendered in virtual machines where input commands are redirected.
While all digital realities are loaded server-side and streamed directly to the user device, 
the physical reality is captured from a (secondary) device placed on a head-mounted display (Figure \ref{fig:headset}), 
where images are sent from the device camera to the server, then processed and loaded in real-time on the \textit{camera feed page} (Figure~\ref{fig:walkthrough}(d)). 
The moment-by-moment changes on a person’s screen or camera feed can be captured as images at a configurable framerate. The captured images are processed and rendered for the user to observe their reality, and also displayed to the user in their view history.
Further, the database loads the corresponding mapping of available interventions (generated by the user, or by the network of users) in the \textit{interventions selection page}. 
The database also loads the view history in the \textit{view history page} (Figure~\ref{fig:tagger}). 
This consists of images of timestamped, visited realities, including chronological images from the camera feed as well as that of the emulators.

Input commands for desktop/mobile 
are captured and directed to the emulator,
including keystrokes (hardware keyboard, on-screen keyboard) and mouse/touch events (scrolling, swiping, pinching, etc).
Screen and camera images are captured at 60~FPS into an \textit{images directory}.
Generated masks and fine-tuned models are stored under an \textit{interventions directory}.
Images and interventions are accessible to their corresponding user.
Interventions are applied sequentially upon an image to return a perturbed image, which then updates the rendered image on the client web app.

\section{Manipulating Realities}
\label{sec:grease}

\subsection{Rendering functionality}

We can decompose rendering functionality into two components:
(i) mapping objects between realities, and
(ii) manipulating the objects within a reality. 
In virtual reality, mapping between real and virtual objects tend to be performed based on relative location in a given view. 
As the digital reality is not a geo-spatial reality, we need an alternative mapping strategy. 
We pursue one where we map based on properties of the objects (e.g. physical appearance/attributes) between realities. 
As such, we built \textit{hooks} to support generalizable identification of object properties between realities. 
We ideate these reality-agnostic hooks based on observed patterns between realities. For example, objects in realities can be encoded in text or graphics, or can be processed by a learnt model if necessary.
Hooks operate time-agnostically:
they can be used to identify object properties during annotation in playback as well as facilitate object manipulation during real-time re-rendering. 
As images are the primary medium of rendering over a reality, the hooks deal with pixel-based inputs and outputs (e.g. though raw text can be extracted from a device emulator, we use OCR to enable generalizable text detection such as in physical realities). 
Overlay modifications are interface-agnostic, enabling the same intervention to scale to all interfaces where the modification condition exists.
It does not require escalation of privilege or modification of source code, and is thus easy to use for users and developers. 

The input receivers for the hooks
begin from the input devices (head-mounted feed, device emulators), 
and into the annotations on the view history. 
The output receivers for the hooks 
begin from the object renders superimposed on the base reality,
and onto the stream of the reality displayed. 
Depending on the objective of the rendering task, how the output of the hooks manifest in the re-rendered reality can vary.
The capability of output renders is independent of the hooks, given that sufficient information or attributes is provided by the hooks.
We focus on \textit{diminished} reality, where we work on reducing visibility on objects.
On the other end of the spectrum, \textit{augmented} reality would be more involved in the addition of new objects or adding/modifying properties of existing objects.

The \textbf{text hook} enables modifying the text on an interface.
Character-level optical character recognition (OCR) takes an image as input and returns a set of characters and their geometric coordinates. 
We first identify a set of regions containing text with EAST text detection~\citep{zhou2017east}. 
We then use Tesseract~\citep{tesseract} to extract characters within each region.
With the availability of real-time textual data from each image instance, an intervention developer can store information processed by the image to be processed by subsequent models via the model hook. 
A sample application of this hook include
interventions against text of specific conditions (e.g. placing censor boxes over hate speech, or generating new text personalized to the user).
Another example is the identification and highlighting of specific text used in one interface (e.g. product ads on Facebook) and appearing in another (e.g. search results in Amazon, or appearing in real-life when in a store).

The \textbf{mask hook} matches the current image against a target template of multiple images.
\textit{Multi-scale multi-template} matching resizes an image multiple times and samples different subimages to compare against each mask instance.
The mask hook can be augmented; for example, using the matching algorithm with contourized images (shapes, colour-independent) or coloured images depending on whether the mask contains (dynamic) objects.
The mask hook could be connected to rendering functions such as highlighting the interface element with warning labels, or image inpainting (fill in the removed element pixels with newly generated pixels from the background).
Given the higher likelihood of non-variability of object instances in the digital world, the applications of this hook would be expectedly predominant in the digital reality. 
An example application is the user can capture a mask of the share buttons on YouTube, and as long as this design is used across all interfaces of YouTube from Android to iOS to browser, this one mask can be reused. Having detected the coordinates of the mask on a given image, the detected object could be occluded, highlighted, or inpainted, etc.

A \textbf{model hook} loads a model to take an input and generate an output.
This enables the embedding of models (i.e. model weights and architectures)
to inform further overlay rendering.
We can connect models trained on specific tasks (e.g. person pose detection, emotion/sentiment analysis) to return output given the image (e.g. bounding box coordinates to filter), and this output can then be passed to a pre-defined rendering function (e.g. draw filtering box).

We enable end-users to tune or adapt their own personalized models
using an annotation interface and model adaptation mechanisms.
Our implementation specifically relies on fine-tuning, but we also review few-shot learning and prompt-tuning. 
For model fine-tuning, the developer re-trains a pre-trained model on a new dataset. This is in contrast to training a model from a random initialization. 
Fine-tuning techniques for pre-trained models, which already contain representations for feature reuse,
have indicated strong performance on downstream tasks \citep{galanti2022on, abnar2022exploring, NEURIPS2020_0607f4c7}.
To retain representations of older tasks or batches of data, online/continual learning methods can assist in reducing catastrophic forgetting \citep{ewc, mota}.
If there is a large number of input distributions
and few samples per distribution,
few-shot learning is an approach where
the developer separately trains a meta-model that learns how to change model parameters with respect to only a few samples. 
Few-shot learning has demonstrated successful test-time adaptation in updating model parameters with respect to limited test-time samples \citep{Raghu2020Rapid, Koch2015SiameseNN, finn2017modelagnostic, datta2021learnweight}.
Some overlapping techniques even exist between few-shot learning and fine-tuning, such as constructing subspaces and optimizing with respect to intrinsic dimensions \citep{aghajanyan-etal-2021-intrinsic, datta2022low, 9157772, https://doi.org/10.48550/arxiv.2205.09891}.
Prompt tuning is an alternative adaptation approach that does not require changes in the parameters of the downstream model.
It is a technique that leverages the use of specific conditioning inputs (e.g. a phrase at the start of a sentence) to condition a foundation model to perform a specific downstream task \citep{lester-etal-2021-power}.
Steps towards scaling the quantity of prompts have been undertaken, 
from PromptSource \citep{bach2022promptsource}, a prompt repository and tool used for creating and sharing prompts, to PromptGen \citep{zhang-etal-2022-promptgen}, a method for dynamic prompt generation.

\subsection{Playback functionality}

Human-in-the-Loop (HITL) learning is the procedure of integrating human knowledge and experience in the augmentation of machine learning models.
It is commonly used to generate new data from humans or annotate existing data by humans.
Examples are reviewed in \citet{https://doi.org/10.48550/arxiv.2108.00941}.

A view history (Figure \ref{fig:tagger}) refers to the historical record (e.g. time series sequence of images) of a view of reality from the perceived viewpoint of the user.
A user can inspect their view history across their camera feed and digital devices, 
and use image segment highlighting techniques to annotate interface patterns to detect and subsequently intervene against these patterns.
The user can go through the sequence of images to reflect on their viewing patterns. 
When the user identifies a GUI element they do not wish to see across interfaces and apps, they highlight the region of the image, and annotate it as \texttt{mask-<name-of-intervention>}, and the mask hook will store a mask of intervention \texttt{<name-of-intervention>}, which will then populate a list of available interventions with this option, and the user can choose to activate it during a session.
When a user identifies text (images) that they do not wish to see of similar variations, they can highlight the text (image) region, and annotate it as \texttt{text-<name-of-intervention>} (\texttt{image-<name-of-intervention>}). The text hook retrieves text, and fine-tunes a pre-trained text classification model on the group of text \texttt{<name-of-intervention>}. For images, the highlighted region will be cropped as input to fine-tune a pre-trained image classification model.

\begin{figure}
    \centering
    \caption{
    \textit{View history page}: The page enables the user to traverse and annotate timestamped images of their view.
    }
    \subfigure[Annotating screen views.]{
    \includegraphics[height=2.7cm]{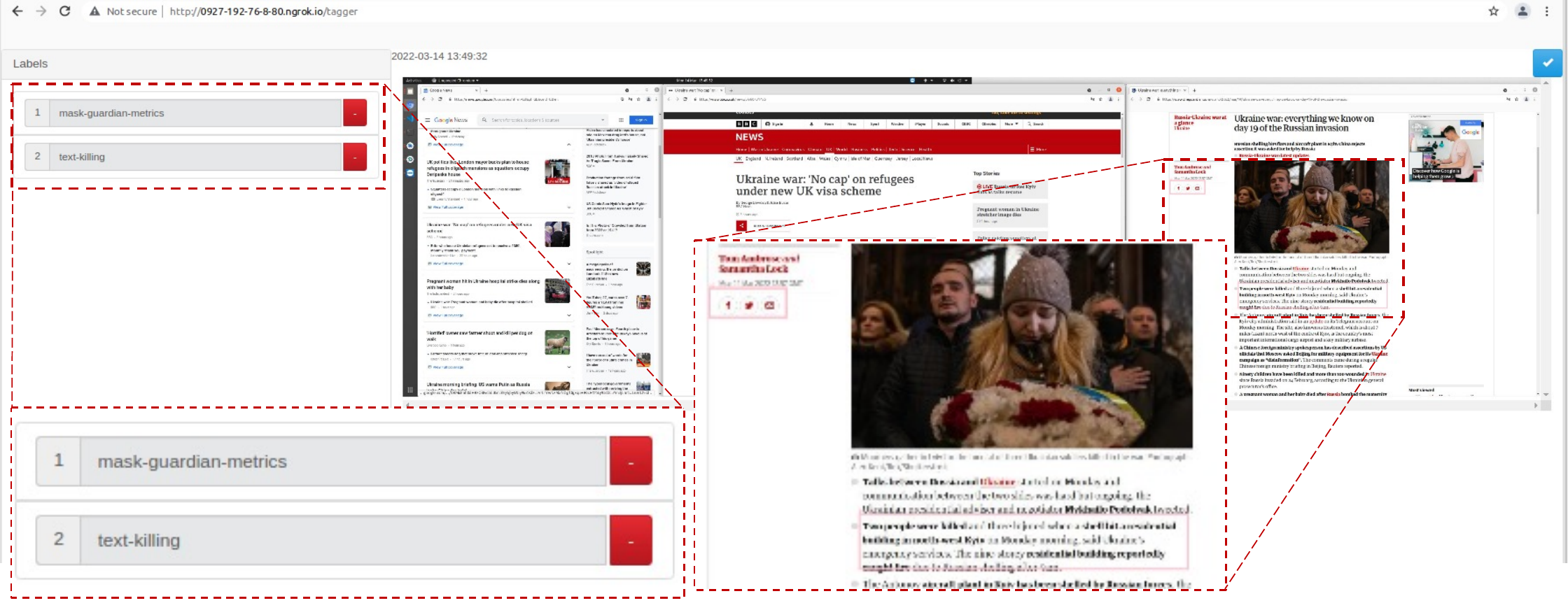}
    }\hfill
    \subfigure[Annotating egocentric views.]{
    \includegraphics[height=2.7cm]{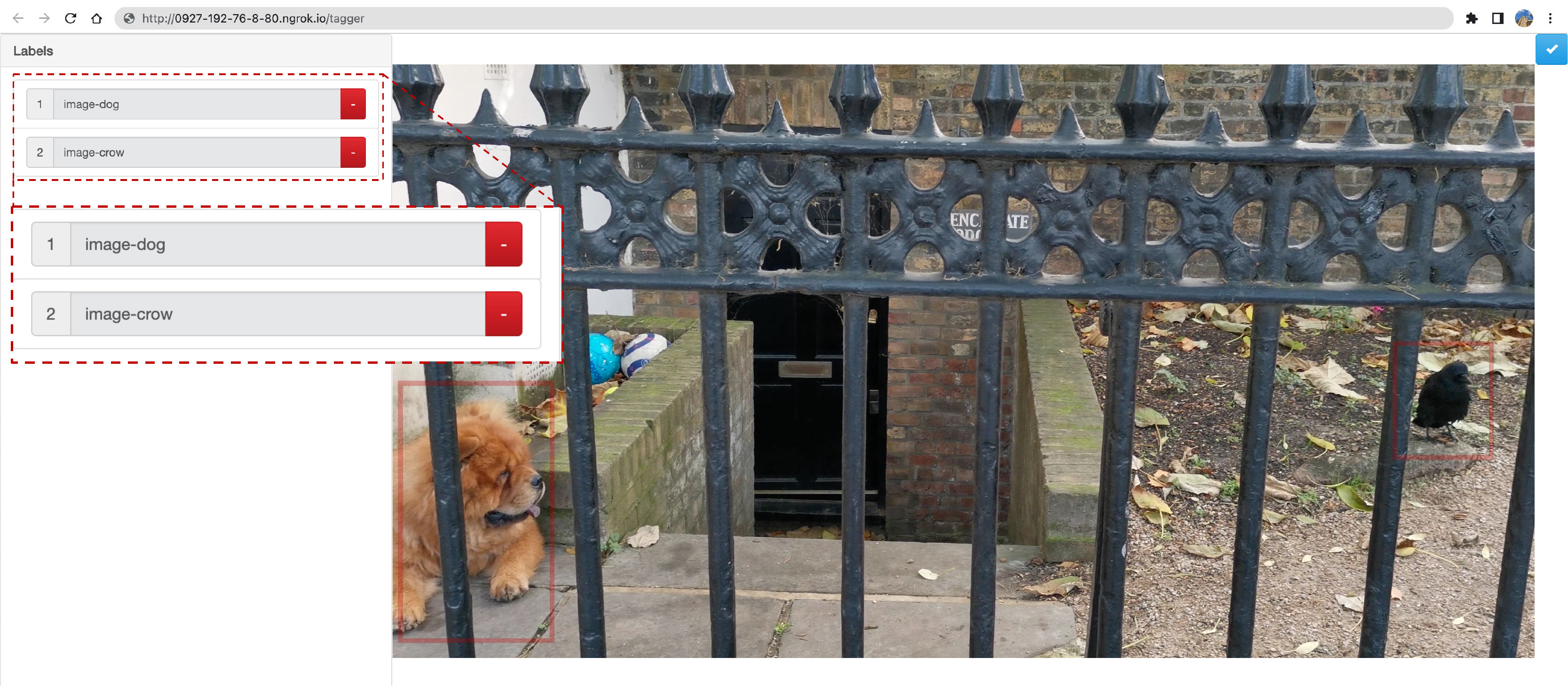}
    }
    \label{fig:tagger}
\end{figure}

\begin{table*}[!ht]
\centering
\begin{minipage}{0.25\textwidth}
    \captionof{figure}{Removal of GUI elements (YouTube sharing metrics) across multiple target interfaces.}
    \subfigure[Desktop (macOS)]{
    \includegraphics[height=2cm]{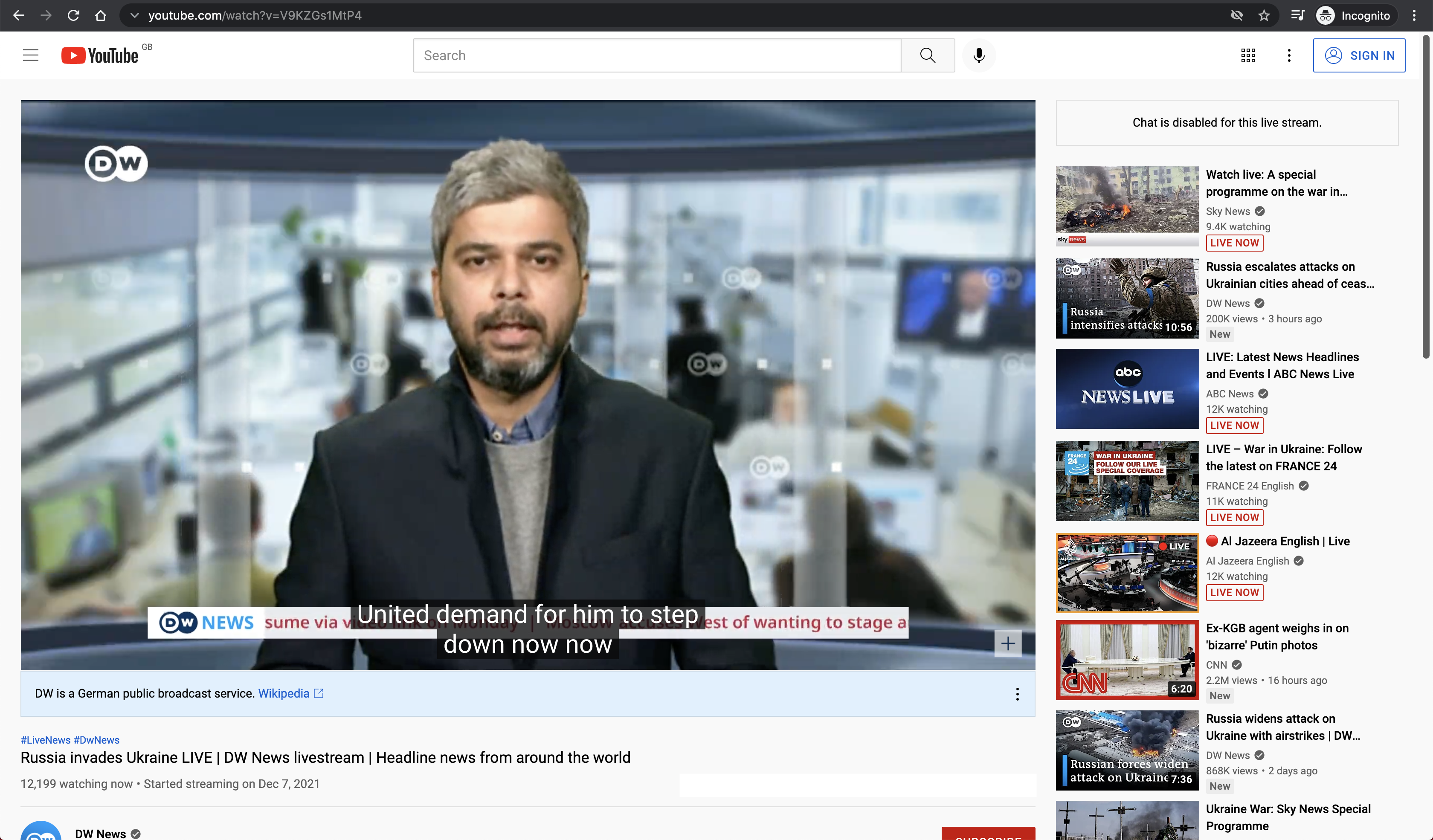}
    } \\
    \hfill
    \subfigure[Android]{
    \includegraphics[height=2.5cm]{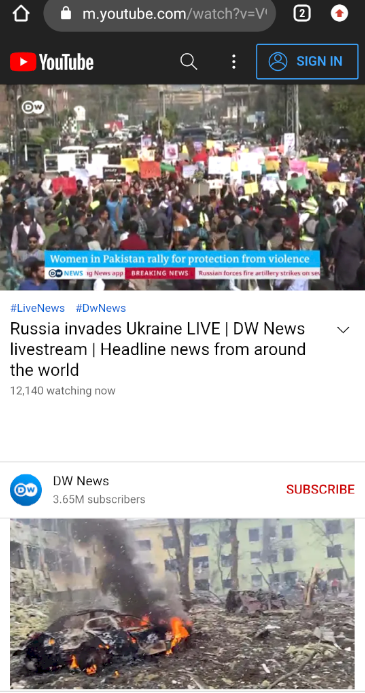}
    }
    \subfigure[iOS]{
    \includegraphics[height=2.5cm]{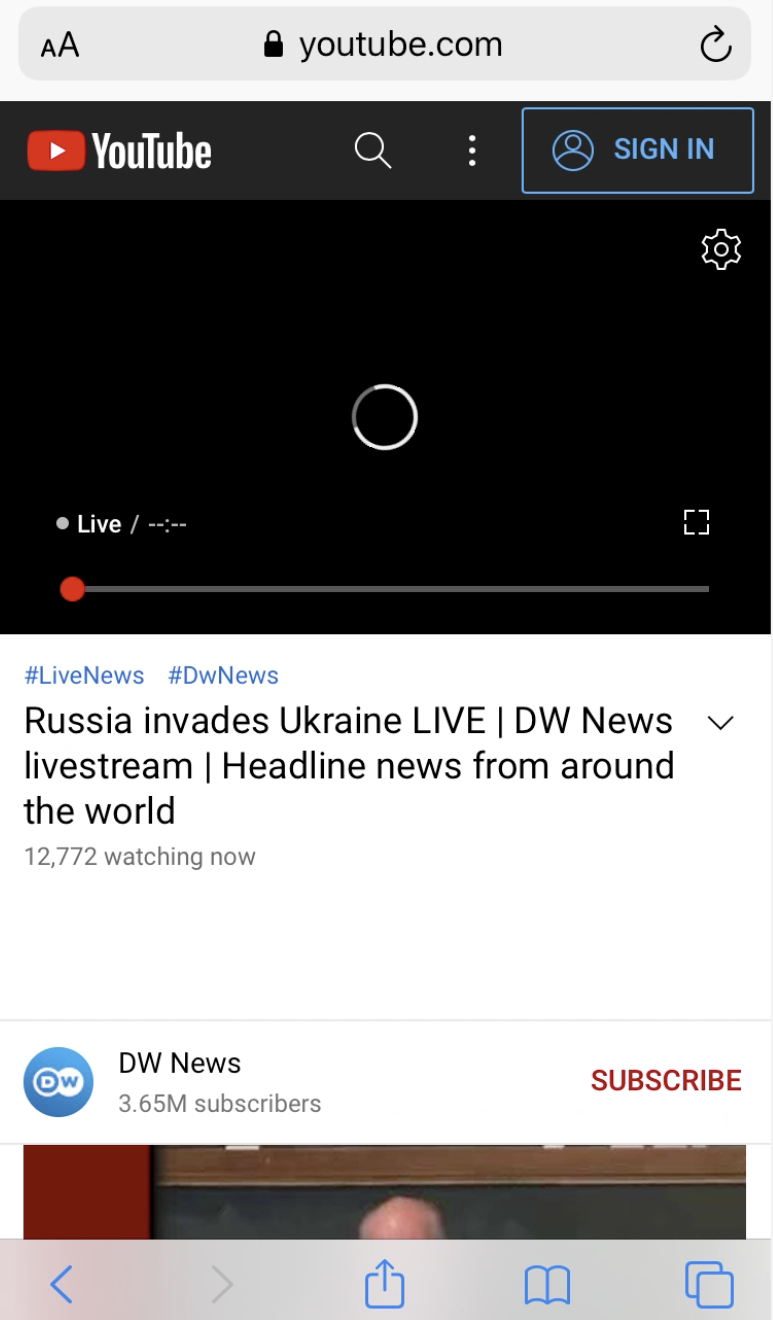}
    }
    \label{fig:sharebuttons_removal}
\end{minipage}
\hspace{0.1cm}
\begin{minipage}{0.725\textwidth}
    \captionof{figure}{Diminished reality in the digital environment (GUI elements, text, images).}
    \subfigure[Occlusion of recommended items (Twitter \textit{top}, Instagram \textit{bottom}) (before \textit{left}, after \textit{right})]{
    \includegraphics[height=4.5cm]{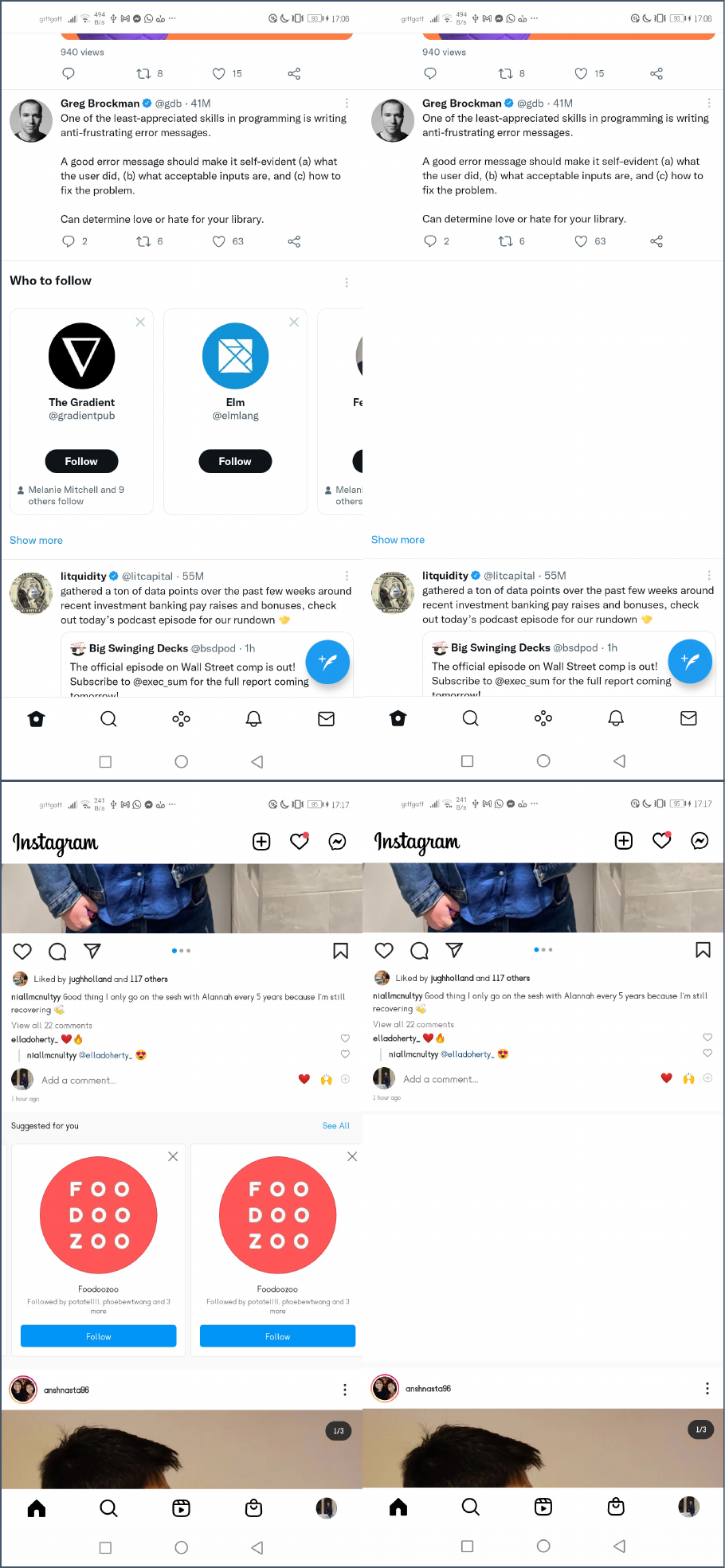}
    }
    \hfill
    \subfigure[Text censoring (YouTube \textit{left}, Reddit \textit{right})]{
    \includegraphics[height=4.5cm]{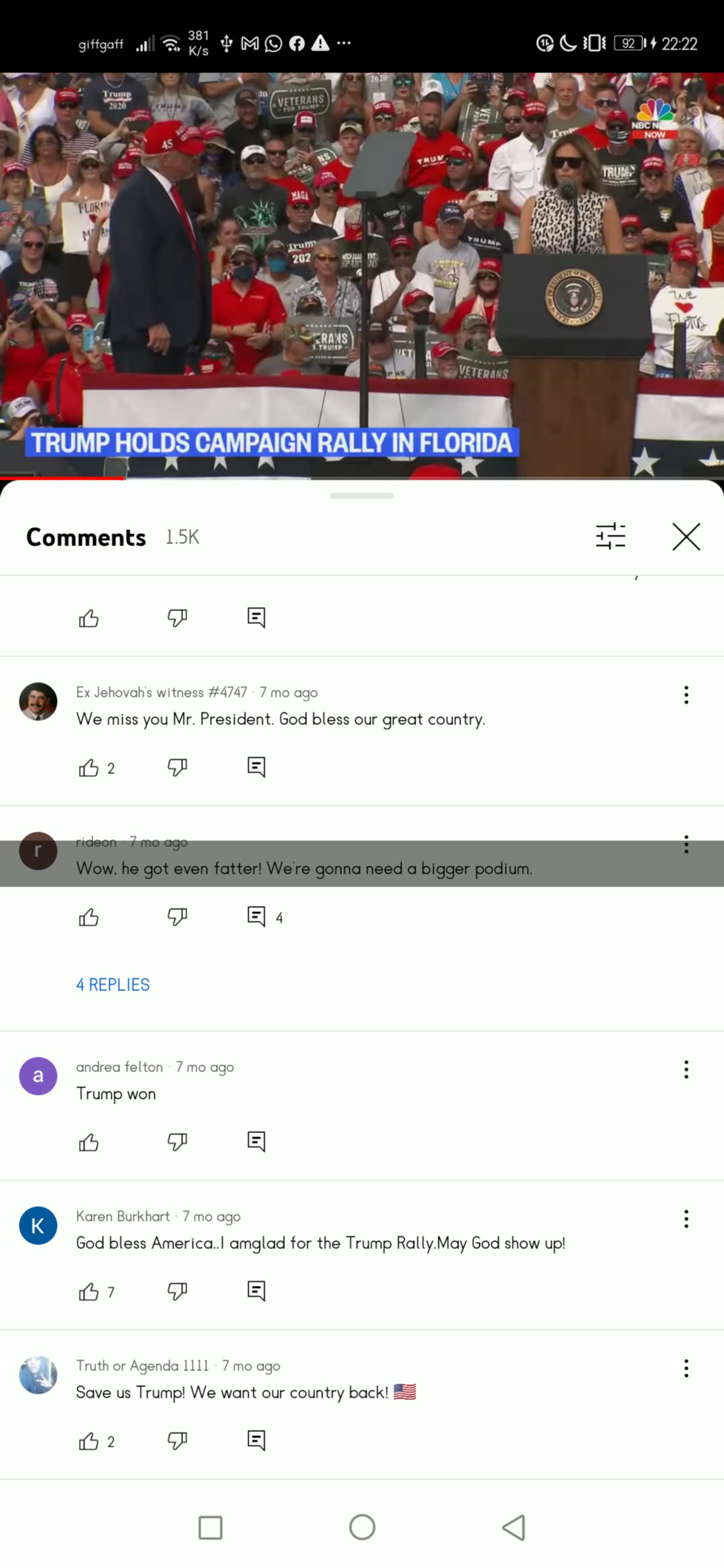}
    \includegraphics[height=4.5cm]{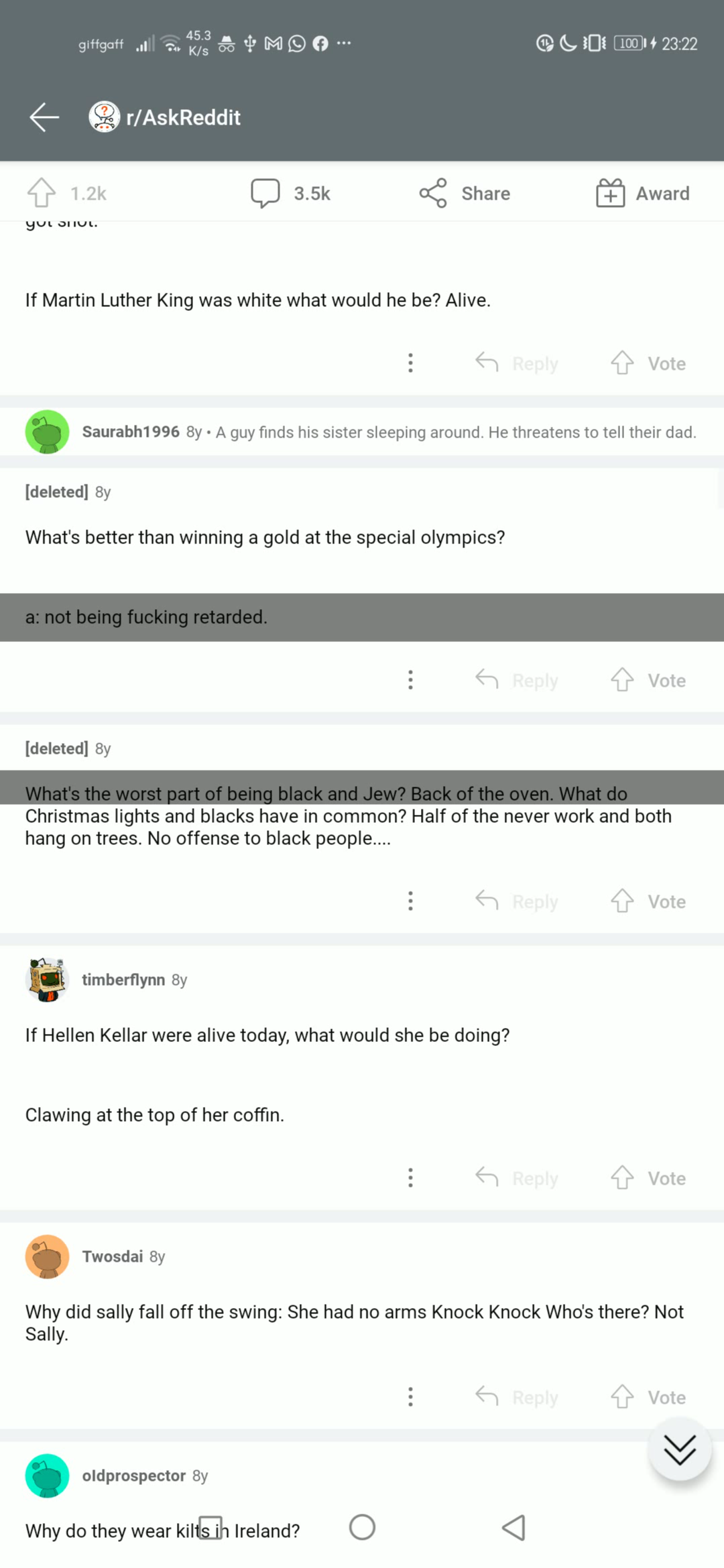}
    }
    \hfill
    \subfigure[Content moderation (Google Images, TikTok, YouTube, YouKu)]{
    \includegraphics[height=4.5cm]{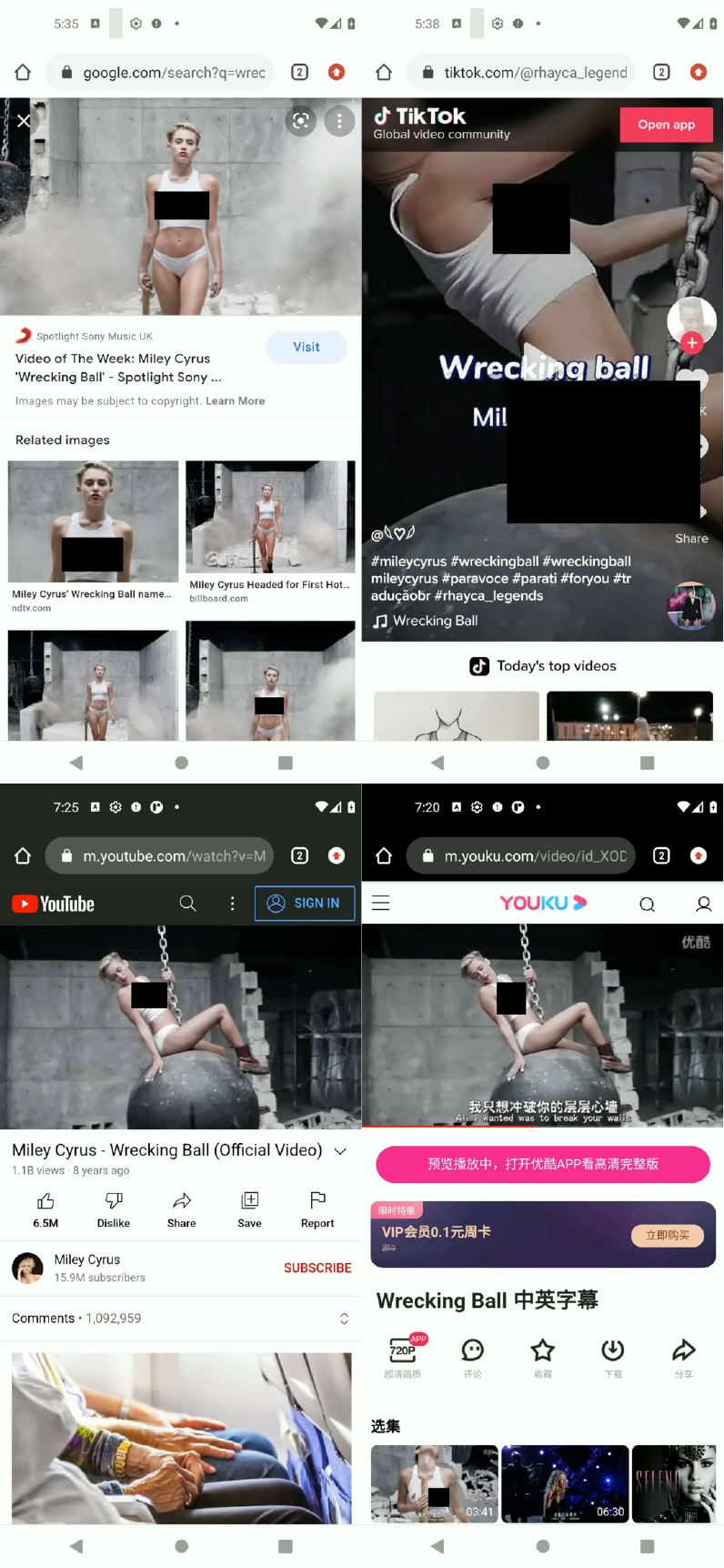}
    }
    \label{fig:gt_eval}
\end{minipage}
\begin{minipage}{\textwidth}
\centering
\captionof{figure}{Time series of occluded egocentric vision from personas evaluation \textit{(chronologically left to right)}.}
\includegraphics[width=\textwidth]{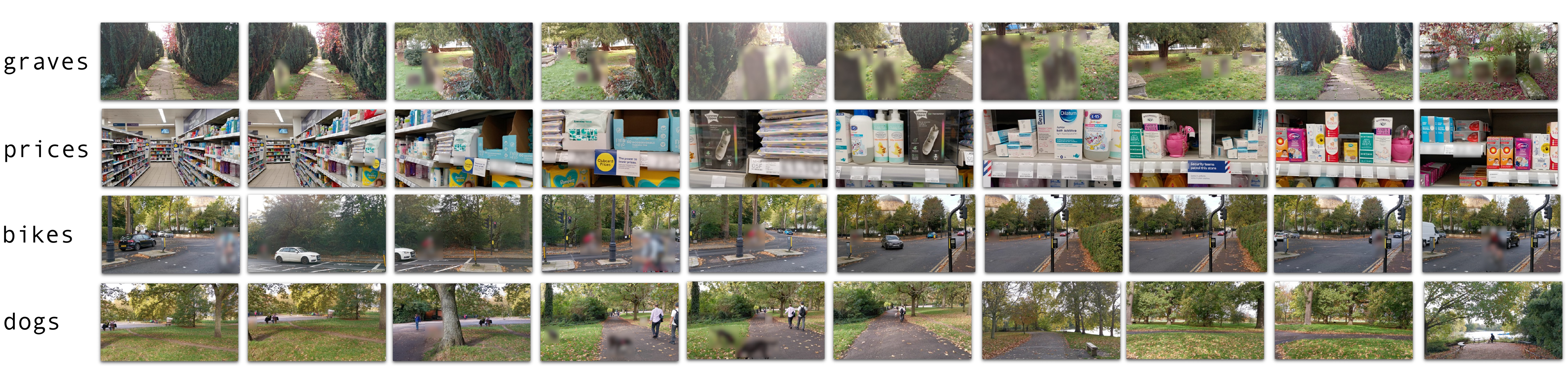}
\label{fig:headset_results}
\end{minipage}
\end{table*}

\section{Evaluation}
\label{sec:evaluation}

\subsection{Cognitive Walkthrough
}

We perform a cognitive walkthrough ("show and tell rather than use and test")~\citep{10.1145/223904.223962, 10.1145/223355.223735}
to simulate a user's cognitive process and explicit actions during usage.
We as the authors presume the role of a user. 
For each step of the walkthrough, 
we first report the data pertaining to each task,
then provide a descriptive evaluation.
To evaluate the process of constructing an intervention,
we track the completion of a set of required tasks (Table \ref{tab:tasks}) based on criteria from
\citet{844354}'s 4 types of automation applications, which aim to measure 
the role of automation in the 
intervention self-development process.
This evaluates the ease and usability in generating interventions and viewing realities.

\begin{quote}{
\begin{description}[noitemsep = 0pt]%
\footnotesize
      \textbf{Step 1: User logs in (Fig. \ref{fig:walkthrough}a)} \\
      \textit{
      The user enters their username/password. These credentials are stored in a database mapped to specific virtual machines that contain the interfaces the user registered for access. 
      }
\end{description}}
\end{quote}

This is a standard step for any secured or personalized system, where a user is informed they are accessing data and information that is tailored for their own usage.

\begin{quote}{
\begin{description}[noitemsep = 0pt]%
\footnotesize
      \textbf{Step 2: User activates interface \& interventions (Fig. \ref{fig:walkthrough}b)} \\
      \textit{
      The user is shown a set of available interventions. They select their target interventions, and select an interface (digital or physical) to view.
      }
\end{description}}
\end{quote}

Users can view the set of interfaces that they can access and use to facilitate their viewing experiences. The interface is available 24/7, retains all their personal data and storage, is recording their view history data for review, and accessible via a web browser from any other device/platform. They are less constrained by the hardware limitations of their personal device.
The populated interventions liken to a marketplace and ecosystem of personalized and shareable interventions. Users can populate interventions that they themselves can generate through the view history tool, or access interventions collaboratively trained and contributed by multiple members in their network. The interventions are also modular enough that users are not restricted to a specific combination of interventions, 
and can apply any combination of interventions sequentially onto the interface.
As the capabilities of generating interventions become extended, so do their ability to personalize their experience, and generate a distribution of experiences to match a similarly wide distribution of users.
The autonomy to deploy interventions, with more options from community contributions, before usage of an interface satisfies Task 4.

\begin{quote}{
\begin{description}[noitemsep = 0pt]%
\footnotesize
      \textbf{Step 3: User accesses the interface and browses (Fig. \ref{fig:walkthrough}c-e)} \\
      \textit{
      The user interacts with the re-rendered interface through their mobile/desktop browser.
      }
\end{description}}
\end{quote}

After the user has chosen their desired interventions,
the user can improve their viewing experience through the manipulation of exposure to certain objects.
The altered viewing experience satisfies both Tasks 1 and 4. Not only is raw image data being collected, but the view is being altered by deployed interventions in real-time.
It is a cyclical loop where users redesign and self-improve their viewing experiences with user-driven tools.

\begin{quote}{
\begin{description}[noitemsep = 0pt]%
\footnotesize
      \textbf{Step 4: User inspects view history to generate interventions (Fig. \ref{fig:tagger})} \\
      \textit{
      After a viewing period, the user may inspect their personal view history to create interventions. They enter the view history page to inspect recorded intervals of their viewing activity across all interfaces, and they can choose to annotate certain regions  to generate interventions.
      }
\end{description}}
\end{quote}

The user is given autonomy in manipulating aspects of their perceived reality.
Enabling the user to inspect their view history across all used  interfaces 
to self-reflect and analyze activity patterns 
satisfies Task 2. 
Though the view history provides the user raw historical data, it may require additional processing (e.g. automated analysis, charts) to avoid information overload.
Rather than waiting for a feedback loop for altruistic developers (e.g. app modifications for digital reality, or dedicated AR software for physical reality) to craft broad-spectrum interventions that may not fit their personal needs, the user can enjoy a personalized loop of crafting and deploying interventions, almost instantly for certain interventions such as element masks. The user can enter metadata pertaining to each annotated object, and not only contribute to their own experience improvement, but also contribute to the improvement of others who may not have encountered the object yet.
By developing interventions based on their analysis, not only for themselves but potentially for other users,
they can achieve Task 3. 

All four tasks, used to determine whether a complete feedback loop between input collection and interface rendering through HITL, 
can be successfully completed.

\subsection{Personas}

Given a set of personalized requirements per user, we evaluate the ability for the system to render interventions. 
We evaluate specifically in physical realities, 
as distributional-diversity of objects are higher in physical realities than digital realities.
Mask hooks work well in highly-uniform realities, such as handling GUI elements in digital realities.
Attributing to the non-uniformity in physical realities, we rely on adaptation with model hooks.

We evaluate using \textit{personas}, which are descriptions of individual people who represent groups of users that would interact with our system.
To construct each persona, 
we populate each hypothetical individual with information pertaining to their background (e.g. context on scenarios or requirements), scenarios (situations/scenarios prompting the persona to use our system), and finally requirements (discrete items that need to be satisfied, generally consistent needs throughout most scenarios).
In-line with \citet{10.1145/2207676.2208573}, 
to evaluate each persona, 
we  
(i) identify persona requirements (i.e. hypothesize scenarios where the persona need our system), and
(ii) evaluate scenario responses (i.e. create real-world scenarios where the personas use the system and evaluate the alleviation of requirements).

Based on the personas constructed in Table \ref{tab:personas},
the visual appearance of the required object tends to be consistent regardless of digital or physical reality.
To support each person's requirements, 
we first manually collected a set of images for each object (\texttt{graves}, \texttt{price labels}, \texttt{bikes}, \texttt{dogs}). 
With our view history tool,
we sample 100 images per object
by searching for each object on Google Images and 
annotating images with object labels and bounding boxes.
Though curated datasets could be found (e.g. dogs in CIFAR10),
we remain consistent in our sampling strategy for objects without datasets.
We fine-tune a Faster R-CNN \citep{https://doi.org/10.48550/arxiv.1506.01497} model, pre-trained on MSCOCO \citep{https://doi.org/10.48550/arxiv.1405.0312}, on each object.
We replace the pre-trained head of the model with a new one containing the new class, and fine-tune until early-stopping at loss 0.1. 
This results in four models for each of the four objects.
With the predicted bounding box coordinates, we apply a Gaussian blur to occlude the object.
After activating these interventions, the authors put on the head-mounted display and enacted the hypothetical scenario in real-life.
We visit physical locations (cemetery, supermarket, street with a cycle path, and dog park), and observe consistent occlusion of the target objects (Figure \ref{fig:headset_results}).
We find that most objects can be successfully occluded in real-time.
Objects that fail to be occluded tend to be at rotated angles inconsistent with sampled images, be a large distance away from the user (i.e. a small set of pixels in the image pertain to the object), or are distributionally-distant from the source distribution of images (e.g. variations of dogs). 

\subsection{Scalability Testing
}

To evaluate the collaborative component,
we measure the improvement to the user experience of a single user from the efforts of multiple users.
We do not recruit 
real users, as it would constrain our performance evaluation to the number of users available, the evaluation period, intervention quality control, and diversity of recruited users. 
Instead, we evaluate through scalability testing, a type of load testing \citep{stest} that measures a system's ability to scale with respect to the number of users.

One application we can use as a base for evaluating our system is the mitigation of different digital harms.
Harms tend to be highly individual and vary in how they manifest within users of digital systems.
The spectrum of harms range from
heavily-biased content (e.g. disinformation, hate speech),
self-harm (e.g. eating disorders, self-cutting, suicide),
cyber crime (e.g. cyber-bullying, harassment,
promotion/recruitment for extreme causes such as terrorism),
to demographic-specific exploitation (e.g. child-unsafe content, social engineering attacks).
We refer the reader to the extensive literature~\citep{hmgov, 10.1145/2998181.2998224, 10.1145/3038912.3052555, 10.1145/3313831.3376370, 10.1145/3359186, https://doi.org/10.48550/arxiv.2210.05791}.
In each subsection, we demonstrate interface-agnostic intervention of different harms specific to GUI elements (Section 5.3.1) and content (Section 5.3.2). 

We simulate the usage of the system
to evaluate the scalable generation of one-shot mask detection, and scalable fine-tuning of text models, 
in order to evaluate the strengths/weaknesses of the system's scalability.
We measure the ease of intervention development with the number of variations of interventions generated (specifically element removal) (Table \ref{tab:mask_results}), rather than development time.
We do not replicate the scalability analysis on real users: the fine-tuning mechanism is still the same, and the main variable (in common) is the sentences highlighted (and their assigned labels and metadata, as well as the quality of the annotations), though error is expectedly higher in the real-world as the data may be sampled differently and of lower annotation quality. 
The primary utility of collaboration to an individual user is the scaled reduction of effort in intervention development. 
We evaluate this in terms of variety of individualized interventions (variations of masks),
and the time saved in constructing a single robust intervention (time needed to construct an accurate model intervention).

\begin{table*}[!ht]
\centering
\begin{minipage}{0.39\linewidth}
\resizebox{\linewidth}{!}{
    \begin{tabular}{p{2.5cm}p{2.5cm}p{2.5cm}}
        \toprule
        \textbf{Task}
        & \textbf{Description}
        & \textbf{Step}	
        \\
        \midrule
        \circled{1} \textit{Information Acquisition} 
        & Could a user collect new data points to be used in intervention crafting?
        & 3 (User accesses the interface and browses)
        \newline
        \newline
        \\
        \midrule
        \circled{2} \textit{Information Analysis}
        & Could a user analyze viewing data to inform them of useful interventions?
        & 4 (User inspects view history to generate interventions)
        \newline
        \newline
        \\
        \midrule
        \circled{3} \textit{Decision \& Action Selection}
        & Could a user act upon the analyzed information about objects they are exposed to, and develop interventions?
        & 4 (User inspects view history to generate interventions)
        \newline
        \newline
        \newline
        \\
        \midrule
        \circled{4} \textit{Action Implementation}
        & Could a user deploy the intervention in future viewing sessions?
        & 2 (User activates interface and interventions), 3 (User accesses the interface and browses)
        \newline
        \\
        \bottomrule
    \end{tabular}
    }
    \captionof{table}{Tasks and the walkthrough steps that satisfy them. 
    }
    \label{tab:tasks}
\end{minipage}
\begin{minipage}{0.6\linewidth}
\resizebox{\linewidth}{!}{
    \begin{tabular}{p{1.5cm}p{5cm}p{5.5cm}p{5cm}}
        \toprule
        \textbf{Persona}
        & \textbf{Background}
        & \textbf{Scenarios}
        & \textbf{Requirements}
        \\
        \midrule
        Persona 1: \textit{graves}
        & 
        They are afraid of death. They went through the trauma of losing a spouse and being forced to quickly bury them in the local cemetery. 
      The sight of gravestones may cause panic attacks.
        & 
        They may pass a graveyard in their local vicinity. Gravestones may also be located in impromptu locations (e.g. points of cycling accidents). Media may contain imagery of deaths, such as in news coverage or leisurely content.
        & 
        Objects, be it in the digital or physical reality, that pertain to death should be occluded from view. 
      An example of such an object is a \textit{gravestone}.
        \\
        \midrule
        Persona 2: \textit{prices}
        & 
        They lack self-control on spending.
      They grew up in poverty and their parents forbade any unnecessary purchases. 
      They purchase anything 'cheap', even if they do not need it. 
      They cannot inhibit their vice when shopping online or in real-life.
        & 
        Priced goods may be shown on e-commerce platforms, advertisements on other webpages or apps, 
      or brick-and-mortar stores.
        & 
        Objects, be it in the digital or physical reality, that display a \textit{price tag} should be occluded. 
        \\
        \midrule
        Persona 3: \textit{bikes}
        & 
        They have anger issues towards cyclists. They witnessed a cyclist run their brother over.
      Now they enter a fit of rage whenever they see one. Not only does this hurt any passing cyclist's feelings who hear their harsh words without any context (and sometimes this escalates into a physical fight), but they also fail to concentrate throughout the rest of the day.
        & 
        They live in a city with extensive cycling routes, resulting in a high prevalence of cyclists.
        & 
        Objects, be it in the digital or physical reality, that pertain to cyclists should be occluded from view. 
      An example of such an object is a \textit{bicycle}.
      Physical safety of the user should be considered. For example, only occluding the bike but not traffic lights, or blurring the bike so that the user still retains depth-perception of an incoming object. 
        \\
        \midrule
        Persona 4: \textit{dogs}
        & 
        They are easily distracted.
      They grew up in an intimately-small family where there were twice as many dogs as there were humans. 
      They now cannot control themselves when they see a dog of any shape or size in real-life. When they see a canine, they stop what they were doing, and start chasing after them. 
        & 
        Dogs can be present on the street, in the park, in the (pet-friendly) workplace, in indoor settings (e.g. cafes/restaurants), etc.
      Targeted advertising tends to show them dog products with their demo dogs. Videos online may also contain dogs. 
      Though they cannot chase them into a screen, the distraction absorbs them and they enter a rabbit hole of browsing funny dog videos. 
        & 
        Objects, be it in the digital or physical reality, that pertain to pets (specifically \textit{dogs}) should be occluded from view. 
        \\
        \bottomrule
    \end{tabular}
    }
    \captionof{table}{Scenarios and requirements evaluated for each persona.}
    \label{tab:personas}
\end{minipage}
\\
\hspace{-1.5cm}
\begin{minipage}{0.49\textwidth}
	\centering
\resizebox{\textwidth}{!}{
    \begin{tabular}{lccccc}
        \toprule
        \textbf{Mask}
        & \textbf{Num. masks}
        & \textbf{Android app}	
        & \textbf{iOS app}	
        & \textbf{Mobile browser}	
        & \textbf{Desktop browser} \\
        \midrule
        \multicolumn{6}{l}{\textit{Stories bar}} \\
        - Twitter & 1 & \ding{51} & \ding{51} & --- & --- \\
        - Linkedin & 1 & \ding{51} & \ding{51} & --- & --- \\
        - Instagram & 1 & \ding{51} & \ding{51} & --- & --- \\
        \midrule
        \multicolumn{6}{l}{\textit{Metrics/Sharing bar}} \\
        - Facebook & 2 & \ding{51} & \ding{51} & \ding{51} & \ding{51} \\
        - Instagram & 2 & \ding{51} & \ding{51} & \ding{51} & \ding{51} \\
        - Twitter & 2 & \ding{51} & \ding{51} & \ding{51} & \ding{51} \\
        - YouTube & 2 & \ding{51} & \ding{51} & \ding{51} & \ding{51} \\
        - TikTok & 2 & \ding{51} & \ding{51} & \ding{51} & \ding{51} \\
        \midrule
        \multicolumn{6}{l}{\textit{Recommended items}} \\
        - Twitter & 2 & \ding{51} & \ding{51} & \ding{51} & \ding{51} \\
        - Facebook & 2 & \ding{51} & \ding{51} & \ding{51} & \ding{51} \\
        \bottomrule
    \end{tabular}
    }
    \captionof{table}{\ding{51} if element removal is successful, \ding{55} if element removal is unsuccessful, --- if the element not available on an interface. 
    }
    \label{tab:mask_results}
\end{minipage}
	\hfill
\begin{minipage}{0.49\textwidth}
        \includegraphics[width=\textwidth]{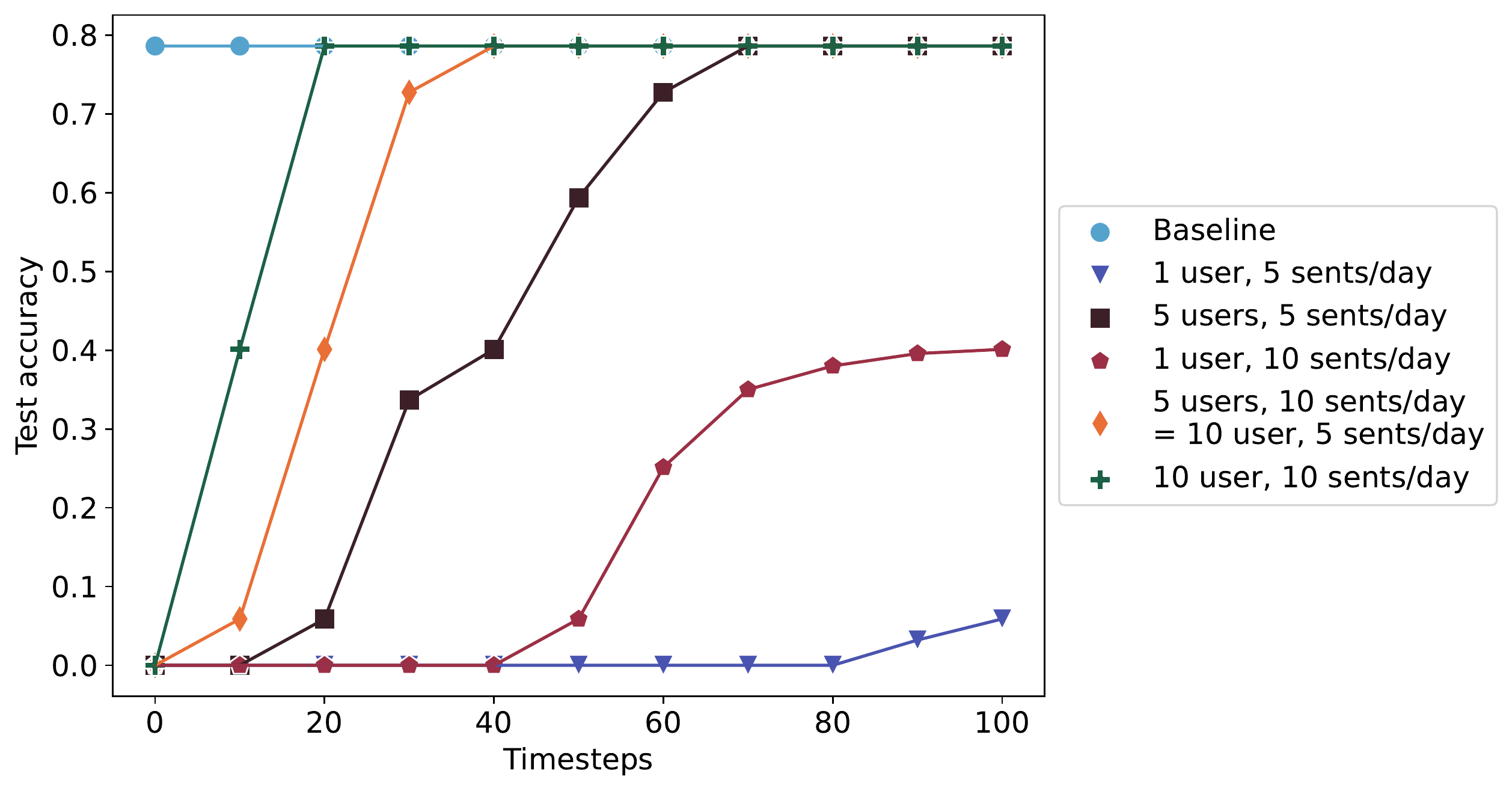}
        \captionof{figure}{Convergence of fine-tuned models on hate speech.}
        \label{fig:fewshot_results} 
	\end{minipage}
\end{table*}

\subsubsection{Scaling mask hooks}

We investigate the ease 
to annotate graphically-consistent GUI elements (Table \ref{tab:mask_results}). 
We sample elements to occlude that can exist across a variety of interfaces.
We evaluate the occlusion of the \textit{stories bar} (predominantly only found on mobile devices, not desktop/browsers).
Some intervention tools exist on Android \citep{swipe, InstaPrefs, gd, greaseterminator} and iOS \citep{friendly}, though the tools are app- (and version-) specific.
We evaluate the occlusion of \textit{like/share metrics}.
There are mainly desktop browser intervention tools \citep{fbd, twd, igd, hidelikes}, and one Android intervention tool \citep{greaseterminator}.
We evaluate the occlusion of \textit{recommendations}.
There are intervention tools that remove varying extents of the interface on browsers (such as the entire newsfeed) \citep{erad, unhook}.
Existing implementations and interest in such interventions indicate some users have overlapping interests in tackling the removal or occlusion of such GUI elements, though the implementations may not exist across all interface platforms, and may not be robust to version changes. 
We evaluate each intervention on a range of target interfaces,
specifically native apps (for Android and iOS) and browsers (Android mobile browser, and Linux desktop browser). 

We use the view history tool to annotate and tag the minimum number of masks needed per element to block across a set of apps. There tends to be small variations in the design of the element between browsers and mobile, hence we tend to require at least 1 mask from each device type. Android and iOS apps tend to have similar enough GUI elements that a single mask can be reused between them. We tabulate in Table \ref{tab:mask_results} the successful generation and real-time occlusion of all evaluated GUI elements. 
We append screenshots of the removal of recommended items from the Twitter and Instagram apps on Android (Figure \ref{fig:gt_eval}a).
We append screenshots of the de-metrification (occlusion of like/share metrics) of YouTube across 
desktop browsers (macOS) and mobile browsers (Android, iOS)
(Figure \ref{fig:sharebuttons_removal}).

\subsubsection{Scaling model hooks}

We investigate the accuracy gains from fine-tuning pre-trained text models as a function of user numbers and annotated sentence contributions (Figure \ref{fig:fewshot_results}). 
Specifically, we evaluate the text censoring of hate speech, where the primary form of mitigation is still community standard guidelines and platform moderation, with little user tooling available on Android \citep{bodyguard, greaseterminator}.
The premise of this empirical evaluation is that we have a group of simulated users $M$ who each contribute $N$ inputs (sentences) of a specific target class (hate speech, specifically against women) per timestep. 
Baselined against a pre-trained model fine-tuned with all sentences against women, we wish to observe how the test accuracy of a model fine-tuned with $M \times N$ sentences varies over time. 
Our source of hate speech for evaluation is the Dynamically Generated Hate Speech Dataset \citep{vidgen-etal-2021-learning}, which contains sentences of \texttt{non-hate} and \texttt{hate} labels, and also classifies hate-labelled data by the target victim of the text (e.g. \texttt{women}, \texttt{muslim}, \texttt{jewish}, \texttt{black}, \texttt{disabled}). As we expect the $M$ users to be labelling a specific niche of hate speech to censor, we specify the subset of hate speech of \texttt{women} (train set count: 1,652; test set count: 187).
We fine-tune RoBERTa \citep{roberta, DBLP:journals/corr/abs-1907-11692}, pre-trained on English corpora Wikipedia \citep{wikidump} and BookCorpus \citep{Zhu_2015_ICCV}.
For each user population $M$ and sentence sampling rate $N$, at each timestep $t$, $M \times N \times t$ sentences are acquired of class \texttt{hate} against target \texttt{women}; there are a total of 1,652 train set sentences under these constraints (i.e. the max number of sentences that can be acquired before it hits the baseline accuracy), and to balance the class distribution, we retain all 15,184 train set \texttt{non-hate} sentences. We evaluate the test accuracy of the fine-tuned model on all 187 test set women-targeted hate speech. 
We also vary $M$ and $N$ to observe sensitivity of these parameters to the convergence towards baseline test accuracy.

The rate of convergence of a fine-tuned model is quicker when the number of users and contributed sentences per timestep both increase, approximately when we reach at least 1,000 sentences for the \texttt{women} category. 
The difference in convergence rates indicate collaborative labelling scales the rate in which text of a specific category can be acquired.
It reduces the burden on a single user of training text classification models from scratch and annotating text alone, diversifies the fine-tune training set, and avoids wasted effort in re-training models already fine-tuned by other users. 

The empirical results from the scalability tests
indicate that the ease of mask generation and model fine-tuning, further catalyzed by performance improvements from more users, 
enable the scalable generation of interventions.

\newpage
\section{Discussion}

\subsection{
The line between virtual reality and physical reality
}

\citet{kishino} elicited the reality–virtuality continuum, 
enabling the interpolation of elements of reality and elements of virtuality. 
They denote the 'real' reality ('reality') as the reality where real objects have objective (physical) existence, 
and the 'virtual' reality ('virtuality') as the reality where virtual objects exist only in essence or effect.
From this, one could conclude that many realities, or views that a user may perceive, are considered as virtual realities.
However, we present some arguments below to revisit some considerations on when a reality is considered 'real' or 'virtual'.

Based on \citet{kishino}'s definition, digital objects do not have an objective existence and would be deemed to be virtual.
However, objects in the digital reality have increasingly real-world effects.
The scope of physical manifestation affects the determination of the virtuality of an object.
Exposure to objects in digital reality or dreams
can affect the cognition of an end-user.
They form new memories and associations, 
and representations/activations in the brain with respect to actions can change over time based on these new experiences.
As such, there is a physical component and manifestation of these 'virtual' objects in the human brain.
Similarly, an e-commerce platform could be interpreted as an input actuator to the transportation of physical objects, even though it does not have the same physical manifestation or even geo-spatial semantic mapping to a brick-and-mortar store.
\citet{skarbez2022revisiting} also revisit the assumptions of the continuum, and find that the 'perfect' virtual reality is unattainable. 
They find that any reality mediated with technology (computers, and extensibly brains) are mixed realities.
They also find that modern virtual reality implementations lie in an interpolated position in the continuum rather than the virtuality endpoint, hence the realities that encompass objects of non-objective existence can be inferred to possess some properties of the physical reality.

The semantic mapping between physical and digital objects are difficult to form, given what ties the physical and digital realities is primarily how the interactions of the user has in each environment have an effect in the real-world (e.g. changes in user cognition, demand-supply of products, voting systems). 
In \citet{kishino}'s continuum, there is an assumed morphism between a pair of realities, 
i.e. a known mapping scheme must exist and be known between a physical and virtual reality. 
However, 
in the digital reality, 
though we can assume that the objects between a physical and digital reality are mappable,
there is no given mapping scheme (instead, we need to learn the scheme), which violates the rules of the reality-virtuality continuum, and thus violating the presumption that the digital world is a virtual reality based on the existence of objects.
Though the objects from this pair of realities cannot be mapped based on affordances with respect to each object (e.g. touching a shoe in a store vs seeing a shoe on Amazon), 
they can be mapped to some extent based on the expectation of the outcome of an action (e.g. owning a pair of shoes upon purchase in a store vs Amazon).

While a pre-defined mapping is not provided between the physical and digital reality, 
as time elapses and the mapping tends towards completeness,
does the status of a reality change accordingly?
We investigated in this paper the potential for end-users to assist cross-reality rendering by providing semantic information collaboratively, 
i.e. using semi-supervised learning to scale semantic mapping.
Given that objects can be interfaced in both the physical and digital reality (unlike a developer-controlled virtual reality) by any number of users publicly,
semantic mapping and labelling can be done collaboratively.
A semi-supervised, specifically human-in-the-loop, approach to constructing mappings between both realities 
helps evaluate strategies towards an unsupervised approach to rendering cross-reality between physical and digital realities, while enabling an initial design to be useful in the near-term.
Enabling end-users to author their own experiences with ScalAR \citep{10.1145/3491102.3517665} have demonstrated improved customization of objects and interactions in custom physical environments. 
screen history
As the learnt mappings between the realities increase over time, in this case through collaborative data collection and intervention generation,
one could argue that any reality can eventually be aligned with respect to the reality-virtuality continuum, and the lack of a pre-defined mapping is not an argument for suggesting the rules of the continuum are violatable.

\begin{figure}
    \centering
    \caption{
    Re-rendering also works in virtual realities. Users can run games on Windows desktops, and the screen history can be annotated. Some of these interventions can also be reused in the digital/physical realities (e.g. if the game is in first-person view). 
    This enables physical/digital reality interventions to transfer over to the virtual reality. 
    It also allows us to augment the interventions by using a virtual reality as a safe annotation environment that approximates the geospatial properties of a physical reality.
    It also allows for easier video game modding without the need for modifying game source code or assets. 
    }
    \subfigure[Grand Theft Auto V: In first-person view, we annotate blood and gore (e.g. from shooting/stabbing).]{
    \includegraphics[width=6.5cm]{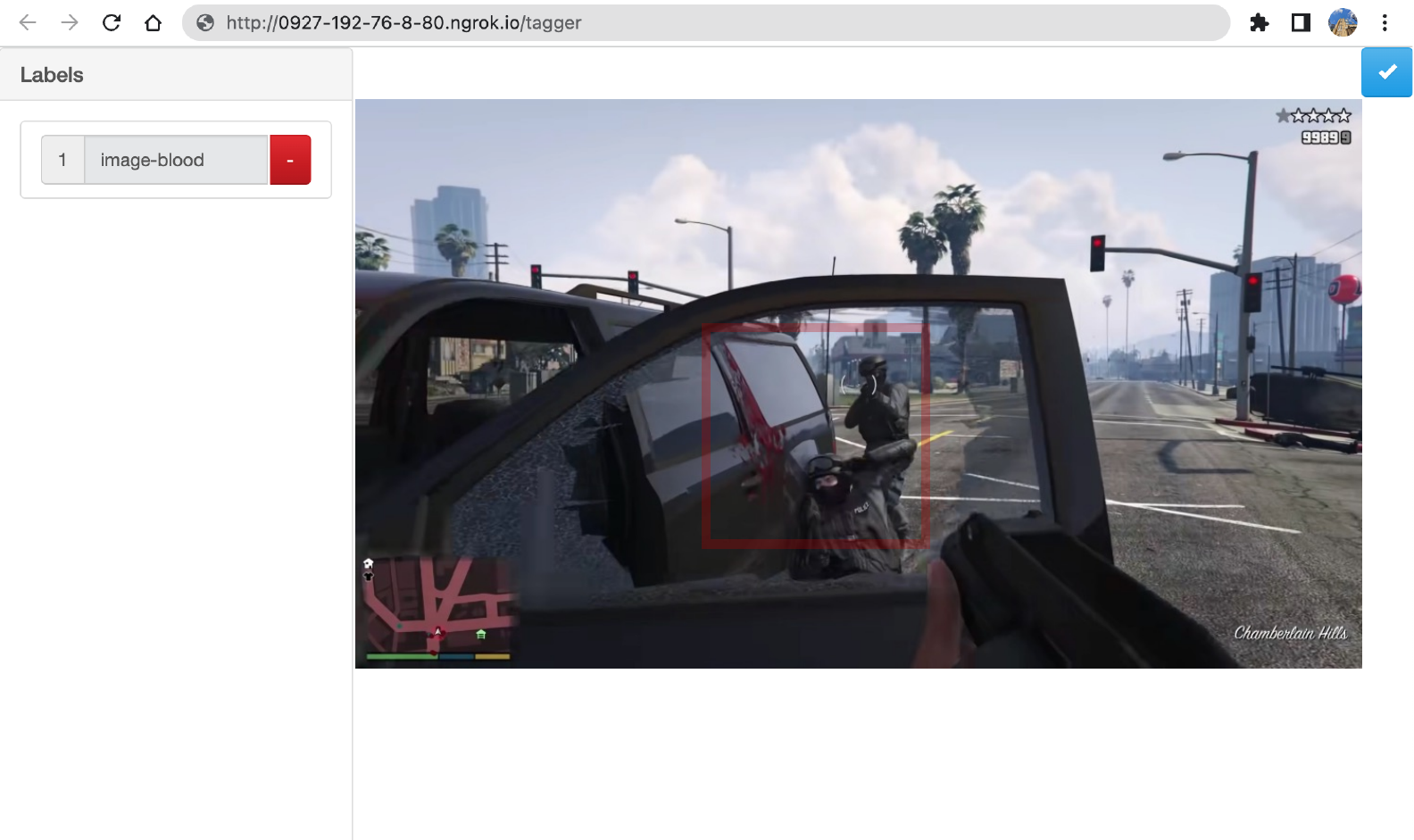}
    }
    \hfill
    \subfigure[Call of Duty Modern Warfare II: In first-person view, we annotate the removal of religious garments, to dissociate the anger towards an enemy with  cultural characteristics.]{
    \includegraphics[width=6.5cm]{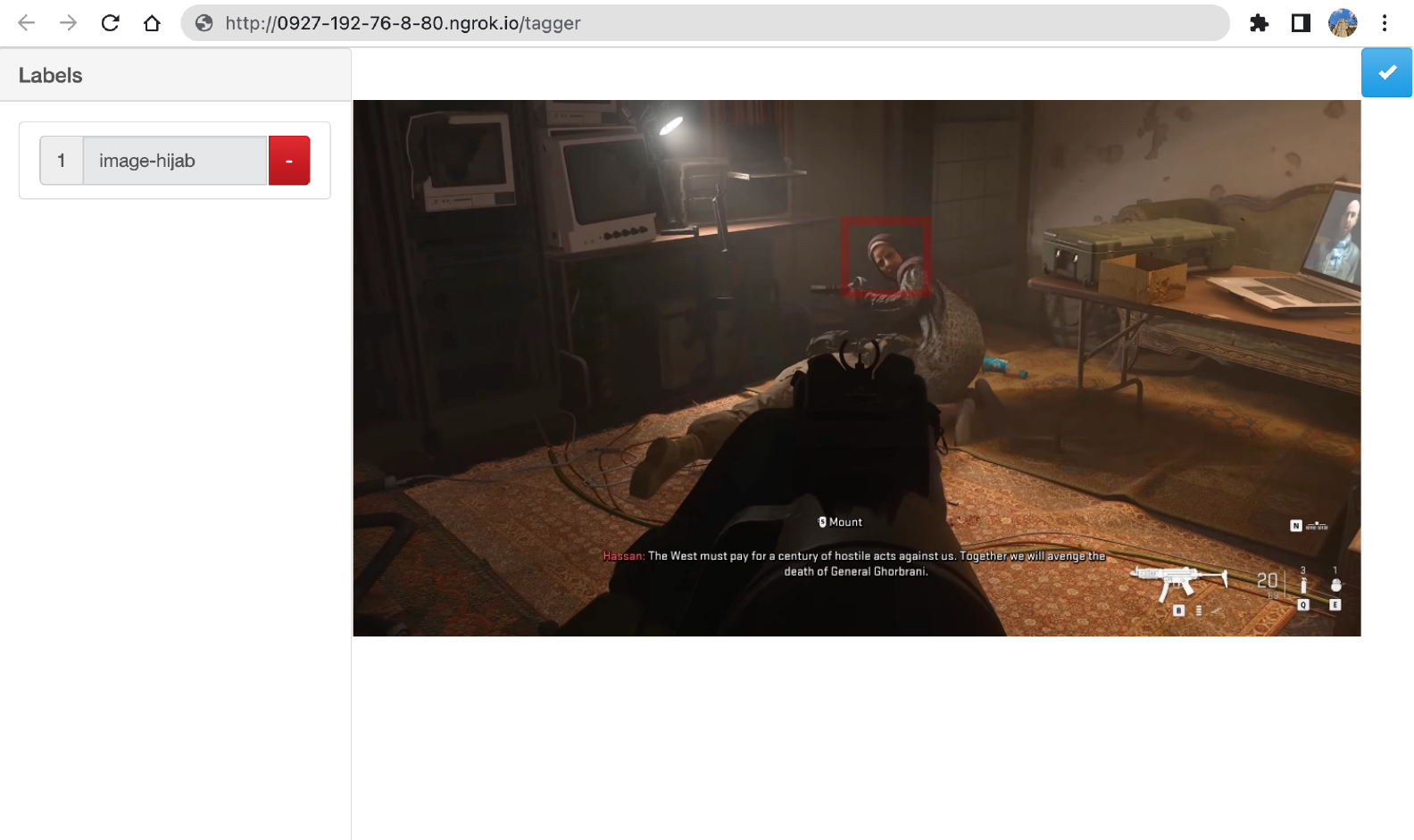}
    }
    \\
    \subfigure[The Witcher 3: In third-person view, we annotate hanged execution.]{
    \includegraphics[width=6.5cm]{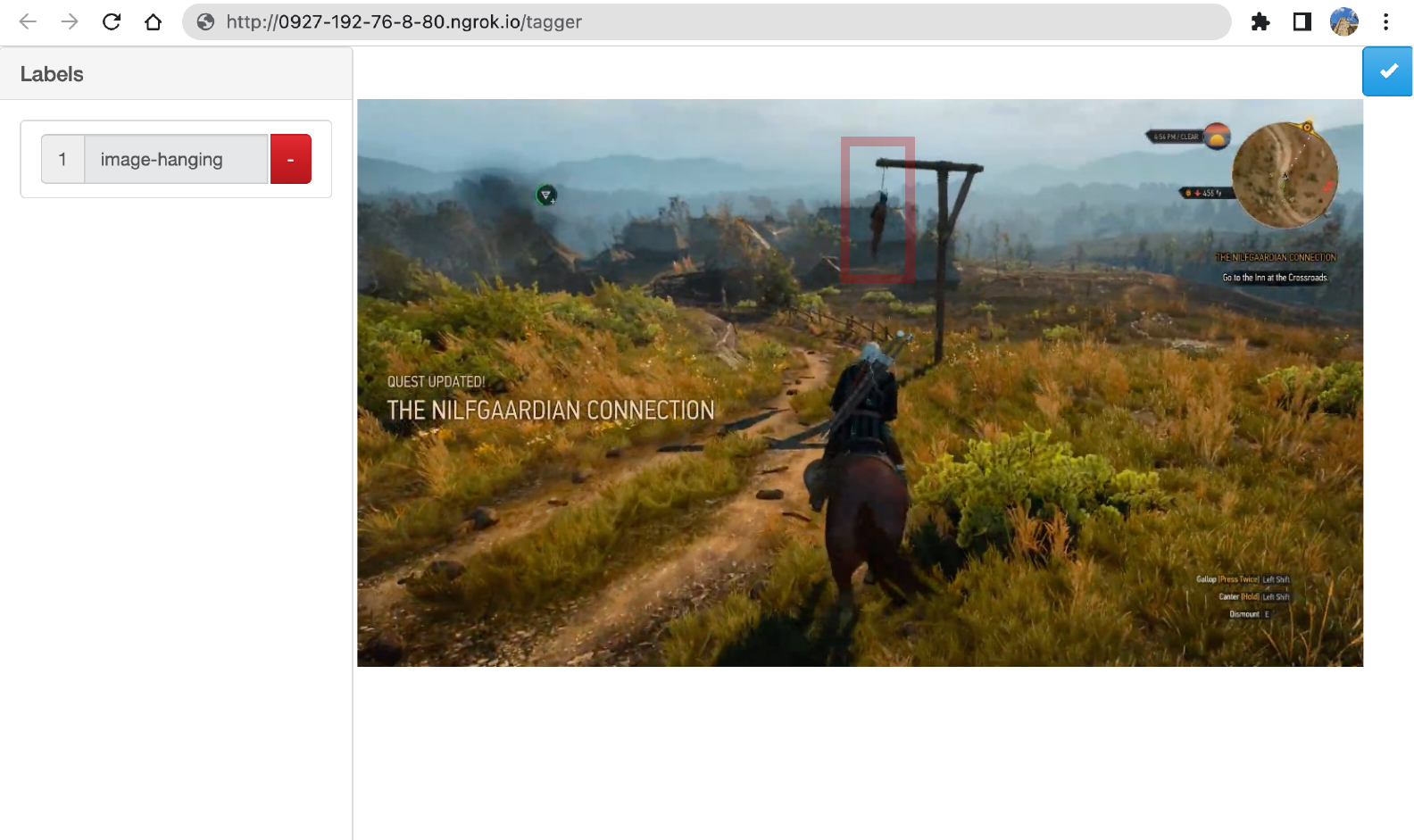}
    }
    \hfill
    \subfigure[Red Dead Redemption 2: In third-person view, we annotate animals hunting.]{
    \includegraphics[width=6.5cm]{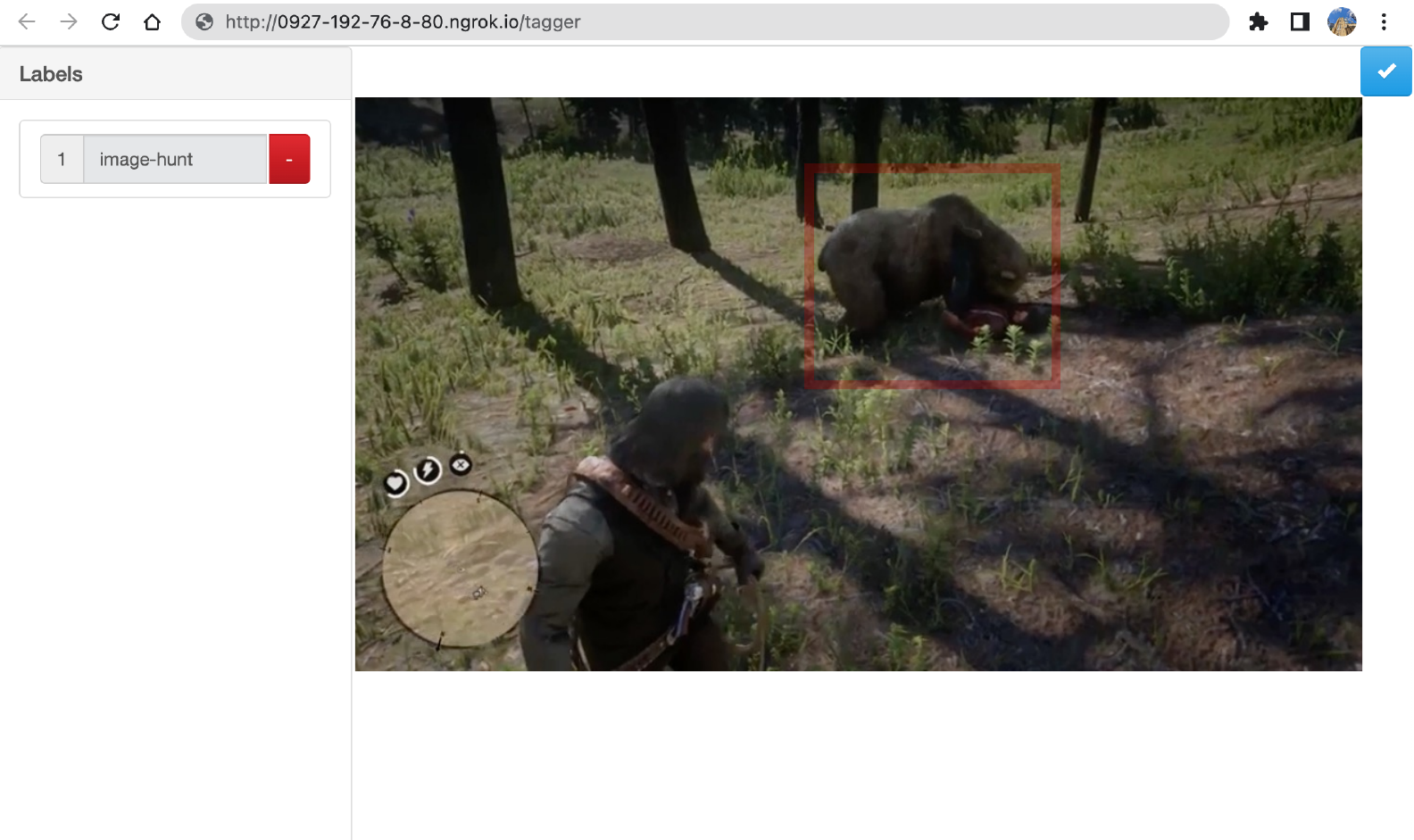}
    }
    \label{fig:tagger}
\end{figure}

In addition to be mappable by effect, objects could be mapped by their interventions.
One can also question how disconnected a physical reality and digital reality is based on the overlap of required interventions.
It could be observed that the issues faced in one reality tend to be consistent in the other.
For example, 
distraction is something a user may take issue with in both physical and digital reality, and the corresponding interventions would be consistent in coccluding the distracting object or changing the lifestyle/habit or interaction flow accordingly.

Virtual realities are also self-contained, in the sense that a specific virtual reality can be rendered with respect to a physical reality, but different virtual realities do not render with respect to other virtual realities. 
Each virtual reality has fixed affordances, and the user usually has had minimal history in each virtual reality. 
Moreover, the instructions needed to modify either reality is distinct.
Traditional virtual realities and the physical reality can be interacted with a similar interaction set (such as gestures, hand movements, touching) with appropriate extensions in the virtual realities.
While physical actions dominate interaction in the physical reality, 
interactions in the digital reality require a different interaction set. Other than user digital actions (such as touching, pinching, typing), developer digital actions exist where modifying source code can modify the digital reality entirely.
There is also a question regarding the level of autonomy that objects possess.
Like real objects, digital objects have an autonomy of their own because they can be tugged around by the rest of the world (e.g. users on a community-driven platform can manipulate digital objects, even if it was originally initialized by an end-user creator). This parallels how a user can own possessions or be surrounded by objects, but they have no objective control over these objects. 
Changing source code is not enough to override its existence in the digital reality; this is unlike virtual reality objects, where changes to them take place primarily with source code changes by a core team of developers.

We also raise the question of perception mediation.
If a user were to wear a cross-reality system for a very long duration of their lifetime's experiences, 
if the visual experience tends towards one where the cross-rendered reality is imperceptible from that of the underlying reality (i.e. the user cannot distinguish that reality is perturbed in any way), 
this perturbed reality could be interpreted as the updated physical reality. 
The introduced objects or properties that do not possess objective existence
do possess existence with respect to the visual perception of the user.
Suggesting that an object that is not present visually or objectively cannot constitute a physical reality
can be compared to
a blind person (or person who has lost all perceptive abilities including sight, hearing and touch) being unsure of an object's objective existence given they cannot visually perceive it.

\subsection{Dreams: Interpolating between real dreams and virtual dreams}

A realm of interest is that of dreams.
Attached to our premise on reality, 
\citet{10.3389/fpsyg.2014.01133} evaluated dreams as a form of simulated reality, termed 'innate virtual reality'.
Similar to a computer-generated virtual reality, 
a dream has limited access to the physical reality, and objects in a dream have no objective existence.
Objects in a dream may not have a physical manifestation, but similar to digital objects, they can also have real-world effects (e.g. through the changes in the user's cognition).
Thus, a key difference between computer-generated virtual realities and brain-generated virtual realities is the 'rendering engine'.
Also similar to our cross-reality rendering premise, 
rendering of dreams can also be manipulated in real-time based on prior conditioning or priming. Users can be primed on certain stimulus, and can subsequently use this stimulus to alter their dreams, such as video or music.
For example, \citet{krakow2010imagery} find that users who visualize a positive end to a nightmare before initiating sleep tends to resolve nightmares during sleep.

We define a \textit{dream} as an environment composed of a sequence of compositionally-generated scenes, 
where the generation occurs through the combination of scenes, scenes' objects, and objects' properties, which originate from a repository of objects/scenes. 
Scenes are a sequence of images from a user's view, that are categorized arbitrarily by the user (e.g. chronological, location-based, action, people).
These viewed scenes are sourced from reality, experienced by the user, and are used as input to render dreams.
From these sourced scenes, the user interpolates properties - rather than interpolating on a continuous space, users interpolate on a discrete space, i.e. enumerating through different combinations of object properties, objects, and scenes. 
As a heuristic as to whether one's dreams are a function of all visual perception, 
\citet{MEAIDI2014586} find that people who are born blind (or become blind early in life) do not experience visual imagery when they dream, and conversely dream with auditory, tactile, gustatory, and olfactory components.
\citet{10.1093/nc/nix001} examined microdreams to find that the content in dream generation is driven by memory, where real perceptions of recent experiences and associated memories form a cohesive image. 

We highlight at least two approaches to rendering dreams: (i) cognitive, and (ii) computational.
Brain cognition is a generative process to render dreams. It can occur consciously (e.g. day-dreams, hallucinations), semi-consciously (e.g. lucid dreams), or unconsciously (e.g. during REM sleep).
We consider generation as an origination problem, where the original composition of objects in a scene is constructed.
Computational rendering of a dream can be both a generative or transcription/translation process. 
Cognitive and computational approaches tend to be used in feedback loops together.
While retrieval has been a common mode of returning unseen scenes using computational approaches, and this retrieval feeds into cognitive rendering (inspiration),
the recent growth in generative models (e.g. DALLE-2 \citep{https://doi.org/10.48550/arxiv.2204.06125}, Imagen \citep{https://doi.org/10.48550/arxiv.2205.11487}, Parti \citep{ https://doi.org/10.48550/arxiv.2206.10789})
permit the sourcing of unseen views of reality and synthesize novel scenes dynamically. 
We use 'unseen' to mean that a user has not experienced, perceived, or viewed a specific instance before.
While retrieval and unconditioned generation are relatively passive modes, 
there are other modes that support more active involvement of the end-user, resulting in human-machine co-creation.
A common computational approach to rendering dreams
is through the use of creativity tools.
When the user has a conscious dream of a scene (e.g. conditioned on a specific task) generated by cognition first,
they then transcribe this dream onto a canvas (e.g. Photoshop for images, or musical instruments for audio, etc).
While generative models are guided by the crowd (crowdsourced datasets), manual creativity tools are guided by the end-user.
The guidance on generation can be a mix of 
experiences of the end-user and outside the end-user. For example, users can provide prompts for conditional generation of outputs. 

As dreams are composed of perceived reality, and only the generation process is affected by the level of consciousness, we do not distinguish unconscious dreams differently from conscious dreams in our evaluation as they are derived from the same inputs, and thus we focus on conscious dreams (and do not propose manipulation of unconscious dreams). 
Some have attempted the manipulation of unconscious dreams, such as Dormio \citep{HAARHOROWITZ2020102938}.

We note some observations about dreams in the context of realities.
We note that a source or repository of realities is always needed to render a dream. 
For example, a source of graphics (Microsoft Clipart, Google Images, icon packs for slide presentations), 
crowdsourced datasets (e.g. for training generative models), 
assets and operations in VR environments, etc.
Though the user has used cognition to generate a template for the dream, it appears users take the rendering shortcut of making use of alternative sources for aiding the final render, and it requires too much manual effort to have a complete render in one's mind to be transcribed onto a computer. 
In addition to stimulation of creativity, such repositories contribute to human-machine co-creation.
Another observation is, similar to reality manipulation, the manipulation of dreams are also through augmentation (adding onto a blank slate) or diminishing (removing objects or tweaking an existing scene). 

For cross-reality systems to assist dreaming, there appears to be a few directions to pursue:
(i) origination (e.g. object/scene generation or retrieval);
(ii) transcription (e.g. interfacing between brain and computer in rendering the dream);
(iii) feedback loop between brain and computer. 
From the perspective of origination, 
the problem is a management of unseen views.
We need to provide novel content that the user has not experienced in their viewed realities.
With the cross-reality system in this paper, we would have a record of all prior viewed realities, so we would know what had not been viewed previously. 
Given the diversity of users and their respective views, we also have a source of unseen realities.
Additionally, we need to consider valid origination. Based on context, we need to know when objects are semantically-valid to be inserted. For example, we cannot just add random objects on any given scene in certain settings. 
From the perspective of transcription,
the problem is a management of seen views.
The goal would be to approximate computationally what is being rendered cognitively with less effort than manual transcription.
Given the user's own source of ideas come from prior viewed realities, we can enumerate through all combinations of objects and scenes in previous views till we obtain the scene the user is thinking of. This is intractable, but it demonstrates an iterative approach to automated transcription, where the user only notifies whether the output matches the cognitive render. 
Another method of minimal-effort interaction is through the use of brain-computer interfaces. For example, \citet{mallett2020pilot} demonstrated a lucid dreamer can control a block on a screen even while asleep.
Prompting is also a common mode of interaction (e.g. in conditional text-to-image generation), where the user provides low-dimensional input (e.g. text) to generate high-dimensional output (e.g. images).
We also need to consider how to manage the feedback loop between cognition and computational rendering.
One part of this is an information visualization problem, as we would wish to avoid information overload for the user. For example, we may wish to make use of "portals" in the regular shapes of some objects, and users can peer into these portals to view the dream. 
Another example is to activate dreaming based on time, place, mental state, or some other conditioning input, similar to how unconscious dreaming is activated when a user falls asleep.
By making use of a specific action or inserting an affordance that is specific to dreams, the user is given the choice and optionality in pursuing a dream, rather than placing dreams everywhere, without filter or choice to not explore if a user already knows they do not like the direction of the dream, etc. 

\section{Extensions \& Near-Future Work}

To extend cross-reality rendering, 
we can consider augmentation in addition to diminishing.
With the progression of new generative diffusion models \citep{https://doi.org/10.48550/arxiv.2204.06125, https://doi.org/10.48550/arxiv.2205.11487, https://doi.org/10.48550/arxiv.2206.10789}, we can explore the use of such models for interface generation in physical and digital settings using different conditioning techniques. 
In terms of object mappings, we currently let the user collect object instances on the view history, and manipulate the object forward in time. We did not manipulate the object in any way backward in time. For example, given the view history of both digital and physical realities, we could use the object mapping to identify patterns of causation (e.g. seeing price tags for footwear products on Amazon and linking it to how a user acts when seeing priced footwear in physical stores). 

To support the scaling of intervention generation and usage, 
we provide users with more data points and pre-populated interventions. 
Automated intervention generation is one approach. Currently intervention generation is a semi-supervised approach, where data is being annotated by users upon self-reflection of specific use cases. We could shift towards automatically generating interventions or recommending the annotation of certain objects to improve the workflow. 
Another approach is to enable an 'other-user' view history mode.
Some tools exist that allow users to see an interface from the perspective of another user (e.g. YouTube \citep{identiswap}). 
If other users feel safe to contribute their view history, they can share it publicly so other users can view it and also annotate it from their unique point-of-view.
Based on user consent, another user can be given another user's egocentric vision to simulate their life, and generate even more interventions based on their personal interpretation.
This can also be pre-populated with the Ego4D dataset \citep{https://doi.org/10.48550/arxiv.2110.07058}, a diverse collection of egocentric vision videos from around the world.
In addition to assisting the mapping in objects between the physical and digital realities, 
hooks assist in scaling the generation of interventions.
Given the limited variability of a specific GUI element on a given app (or ease of re-cropping an updated interface design), one shot of a GUI element is sufficient to detect it. 
As such, mask hooks tend to require only a single cropped image as input, and we sample a large number of GUI element interventions based on existing digital/perceptual harms literature.
Pre-populated model hooks can also be sourced from model sharing platforms (e.g. AdapterHub, huggingface, PapersWithCode, Github, ModelZoo).

Other than benefiting from existing contributions in the machine learning ecosystem, 
this system can also contribute back to the same machine learning ecosystem.
This system can contribute to the pool of task-specific fine-tuned models,
annotated/labelled datasets on various tasks, 
datasets of high distributional shift (attributed to the non-uniformity of user experiences),
or providing unlabelled view history data for unsupervised tasks.
This helps developers working on reality manipulation (e.g. AR/MR/VR researchers, digital harms researchers)
by providing them with data on what users wish to mitigate (e.g. a repository of digital harms), or
initial user-initiated designs on what interface changes users would like to see (and thus be implemented natively).

\section{Limitations}

The current system design is the product of numerous iterations. In each iteration we aimed to resolve different challenges that posed as hurdles to usability and deployment. 
We opt for a complete server-side implementation, where we load devices and run interventions on a server, and stream to client devices.
This reduces the burden on end-users for client-side hardware specifications.
With our setup, the user does not need a high-end smartphone or specialized AR/MR/VR hardware with built-in processors. 
A user can use a smartphone of any specification (as long as it has an internet connection and load webpages) to load any device emulator. 
They can procure a headset for mounting a smartphone, 
and this can be as costless as building one out of cardboard \citep{cardboard}.
In a previous iteration, we attempted to keep the device interface on the client-side and stream just-in-time overlay renders. While this worked in most settings given sufficient bandwidth, our concern lied in the off-chance that a user with insufficient bandwidth might see an overlay render after the underlying interface image had changed.
We concluded it was better to liken the access to manipulated realities to that of buffering a video; if a user prioritizes interventions, they may be willing to stream an interface (even with rare delays), and video buffer time has been drastically shortened over the years with improvements in streaming architectures and bandwidth access.
We also move away from a code modification approach to changing interface functionality.
By identifying commonalities between interfaces agnostic to operating system (native program patches tend to be OS-specific), 
we allow objects to be manipulated across operating systems.
Not all interfaces have an underlying 'code' that can be modified, such as the physical reality, and thus overlays have been the predominant strategy in modifying physical realities. 
To maintain a reality-agnostic approach to manipulating realities, 
we adhere to the use of overlays.
Overlays make use of what the user can see as their input and output.
Further, prior use cases for reality manipulation tend to require a third-party to craft interventions for the user, 
be it patch developers for app modifications,
or developers for AR/MR/VR software.
From a development cycle where 
users and developers engage in a feedback loop 
to maintain and upgrade software over time,
we directly support users in maintaining or developing software themselves.
They can craft their own interventions that perform specific digital functionality or AR/MR/VR functions.
Despite this progress, 
there are still avenues for improvement.
We highlight some extensions needed to improve the overall user experience, 
as well as preliminary directions on how to approach the limitations.
Most of the following limitations are not critical issues with the system design;
conversely, they are at most 'band-aids' that can be plastered onto the system to improve the experience, but they do not break the experience.

Some failure modes are component-specific; they are not a failure in system design necessarily, but requiring improving individual components.
The mask hook might face difficulty in element removal of ‘dynamic’ elements (e.g. removing the video box for YouTube videos if we exclude the sharing metrics), or the overblocking of elements (e.g. removing the homescreen).

Our current implementation has handled most interaction modalities pertaining to imagery and text, 
but there are other modalities that would need to be manually built.
Accessing client-side hardware is possible (e.g. \citet{vx}), such that the server-side emulator can access the user's local camera, audio speakers, sensors, and haptic vibrators. 
Furthermore, while we have provided hooks for image-based interactions (e.g. text can manifest as an image in any reality), 
we did not implement a hook for audio/speech.

Safety is also an important concern. As a user can manipulate their physical realities, there may be some critical situations where the re-render needs to be undone, or the user should be informed of the non-overlayed reality. For example, when crossing the road, though bicycles are occluded, they should not be completely inpainted. They should be slightly blurred, or at least a big arrow should above the cyclist to inform the user that an object exists and is approaching them. This also means certain objects that are intended to be used for physical safety, such as fire extinguishers or traffic lights, should not be overlay-able. We could insert safety checker models to verify that non-overlayable objects are not manipulated, or alternatively we could prompt the user to re-consider their decision (e.g. doing a sample playback in the view history of what happens when this object is occluded).

There are a few considerations regarding scaling, in terms of model development and data quality.
An assumption made is that users in a network know a ground-truth label of the category of the specific text they wish to detect and occlude, and the crowd-sourced text of each of $N$ categories will yield corresponding $N$ fine-tuned models.
A concern with this assumption, is that the labelling of such inputs in the real-world may not be standardized, and similar inputs may be grouped separately or dissimilar inputs may be grouped together, if we purely rely on network-based tagging. 
We may encounter scenarios of 
out-of-distribution shift (e.g. users sample non-uniform sentences), adversarial samples (e.g. users maliciously tag sentences that worsen accuracy).
On the one hand, we can evaluate a tagging system that shows the user similar intervention tags as the one they are entering, so that an existing intervention is updated rather than creating duplicate interventions.
On the other hand, perhaps the data points are indeed distributionally different from an existing tag's dataset, so creating a different tag would be appropriate.
Possible algorithmic approaches to ensuring similar texts are grouped together for fine-tuning could be the use of in/out-of-distribution detection (e.g. computing the interference in loss convergence with respect to 2 inputs coming from different categories, or using a similarity metric, in order to regroup contributed inputs into appropriate categories), or the use of ensemble models (e.g. preparing $M$ different batches of training sets to train $M$ different ensemble models, so that the dissimilarity between certain sentences do not afflict a single model alone, and other models can validate a prediction).
Furthermore, to reduce the reliance on the user population and sampling rate for intervention generation, we can explore faster adaptation methods (e.g. batch-efficient fine/prompt-tuning with large foundation models) so that less/no additional data would be needed for generating interventions. 
Alternatively, we can pre-populate the system internally first with a large number of masks and models (e.g. sourcing models from AdapterHub, github, modelzoo, etc).

\section{Conclusion}
\label{sec:conclusion}

Our cross-reality re-rendering system
supports users in manipulating their digital and physical realities. 
They can inspect their historical views, annotate objects, and share the interventions they generate.
We evaluate requirements  
with cognitive walkthroughs, scalability tests, and personas.
We hope this work continues to inspire further exploration of cross-reality systems into other realities.

\newpage
{\small
\bibliography{main}
\bibliographystyle{ACM-Reference-Format}
}

\end{document}